%% file: acl_latex.tex
\documentclass[11pt]{article}

\usepackage[final]{acl}

\usepackage{times}
\usepackage{latexsym}
\usepackage{amsmath}
\usepackage{booktabs}
\usepackage{multirow}
\usepackage[T1]{fontenc}

\usepackage[utf8]{inputenc}

\usepackage{microtype}

\usepackage{inconsolata}

\usepackage{graphicx}
\usepackage{subcaption}
\usepackage{tabularx}
\usepackage[table]{xcolor}

\title{When Knowledge Is Not Free: Cost-Aware Evidence Selection in Retrieval-Augmented Generation}

\author{Mingyan Wu$^{1}$\footnotemark[1], Han Yang$^{2, 3}$\thanks{\ \ indicates equal contribution.}, Omer Ben-Porat$^{4}$, Yftah Ziser$^{5,6}$\thanks{ \ \ indicates corresponding author.} \\ 
$^1$Northeastern University, 
$^2$ Technical University of Munich,\\
$^3$ GESIS – Leibniz Institute for the Social Sciences,\\
$^4$ Technion--Israel Institute of Technology,
$^5$ NVIDIA Research,
$^6$ University of Groningen
}

\begin{document}
\maketitle
\begin{abstract}
Retrieval-Augmented Generation (RAG) typically assumes that external knowledge
is free, but many high-quality sources are paywalled, licensed, restricted, or
otherwise costly to access. We introduce cost-aware RAG, a setting where
retrieved evidence is assigned access-cost tiers and systems must answer under
an explicit evidence-access budget. We instantiate this setting by augmenting
MS MARCO v2.1 with access-friction tiers and evaluate budgeted evidence
selection across general-domain and domain-specific QA benchmarks. Our results
show that static selection is brittle: no fixed selector uniformly dominates,
and larger budgets do not reliably improve answer quality, even when costly
evidence is domain-matched. We then study agentic cost-aware RAG, where an LLM
decides when to retrieve, which tier to access, and when to stop. Agents show
strong promise as adaptive evidence-acquisition controllers, but their behavior
remains highly model- and task-dependent. These findings suggest that
cost-aware evidence acquisition is a central challenge for the next generation
of RAG systems. 
All code and data are available at \url{https://github.com/Mignonmy/Cost-Aware}.
\end{abstract}

\input{latex/0_abstract}
\input{latex/1_introduction}

 \input{latex/2_data_construction}

 \input{latex/3_methodology}
\input{latex/4_results}
\input{latex/5_agent}
\input{latex/7_related_works}
\input{latex/5_conclusion}

\input{latex/6_limitation}
\section*{Acknowledgement}
Han Yang received funding from the Deutsche Forschungsgemeinschaft (DFG) under grant number: MA 3964/15-3 (SocioHub project). Han Yang received additional funding from the European Union under the Horizon Europe grant OMINO -- Overcoming Multilevel INformation Overload\footnote{\url{https://ominoproject.eu/}} under grant number 101086321 \cite{informationoverload}.

\bibliography{custom}
\clearpage
\input{latex/8_appendix}

\end{document}

%% file: latex/1_introduction.tex
\section{Introduction}

Retrieval-Augmented Generation (RAG) systems are usually evaluated as if external knowledge were freely and uniformly accessible. In practice, however, many high-quality sources are restricted, licensed, paywalled, or otherwise costly to access, including academic journals, clinical references, professional databases, licensed code repositories, and proprietary business resources. This makes retrieval not only a relevance-ranking problem, but also a resource-allocation problem: a system must decide which evidence is worth acquiring under a limited access budget. This access cost is distinct from commonly studied RAG constraints such as computation time, context length, or token budgets \cite{jeong2024adaptive,yue2025inference,lu-etal-2025-turborag}. Those constraints limit how much evidence can be processed once it is available; access costs determine which evidence can be obtained in the first place. Cost-aware RAG therefore requires estimating the value of evidence relative to its cost, rather than optimizing retrieval relevance or generation quality alone.

\input{figure_tex/intro_fig}

We study this setting by introducing a cost-aware RAG evaluation framework that augments MS MARCO v2.1 with discrete access-cost tiers. These tiers serve as coarse proxies for access friction, distinguishing freely accessible sources, general open-web sources, and higher-friction professional or restricted sources. As shown in Figure~\ref{fig:cost-aware-rag-setup}, each evidence item has an access cost, and the system must select evidence under an explicit budget before passing it to the generator. We evaluate this framework on general-domain QA benchmarks, including HotpotQA, Natural Questions (NQ), and TriviaQA, and domain-specific medical QA benchmarks, including MedQA-US and MMLU-Med. For the medical setting, we augment the high-cost tier with domain-matched textbook evidence, allowing us to test whether costly specialized sources help when relevant knowledge is concentrated in higher-friction resources. We compare static evidence-selection strategies, including cost-agnostic top-\(k\), relevance-only selection, cost-aware greedy selection, knapsack-based optimization, and redundancy-aware methods.

Our results show that cost-aware RAG does not admit a single fixed selection rule. No static selector uniformly dominates across datasets, budgets, and backbone models, and larger access budgets do not reliably improve answer quality. Since the generator receives a bounded evidence context, increasing the budget mainly changes the composition of the selected evidence, allowing low-cost passages to be replaced by higher-cost ones. Even in the medical setting, where costly evidence is domain-matched, a more expensive evidence mix is not necessarily more useful. These findings motivate an adaptive formulation. Rather than assigning every query the same budget and applying a fixed selector, a cost-aware system should decide during inference whether more evidence is needed, whether higher-cost evidence is worth acquiring, and when the current context is sufficient. We therefore study an agentic formulation of cost-aware RAG, where an LLM acts as an evidence-acquisition controller. Agents can substantially improve the performance--cost tradeoff with some backbones, but their behavior remains model- and task-dependent, showing that current LLMs are still imperfect cost-sensitive evidence controllers.

\paragraph{Contributions.}
We make three contributions:
(i) We introduce a cost-aware RAG evaluation setting by augmenting MS MARCO v2.1 with discrete access-cost tiers that model relative access friction.
(ii) We benchmark static budget-constrained evidence-selection strategies across general-domain and domain-specific QA benchmarks, showing that no fixed selector uniformly dominates and that larger access budgets do not reliably improve answer quality.
(iii) We study agentic cost-aware evidence acquisition in per-query and shared-budget settings, showing that agentic frameworks are promising for cost-aware RAG while revealing that current LLMs remain unreliable cost-sensitive evidence controllers.

%% file: figure_tex/intro_fig.tex
\begin{figure}[t]
    \centering
    \scalebox{1}[0.90]{%
        \includegraphics[
            width=\columnwidth
        ]{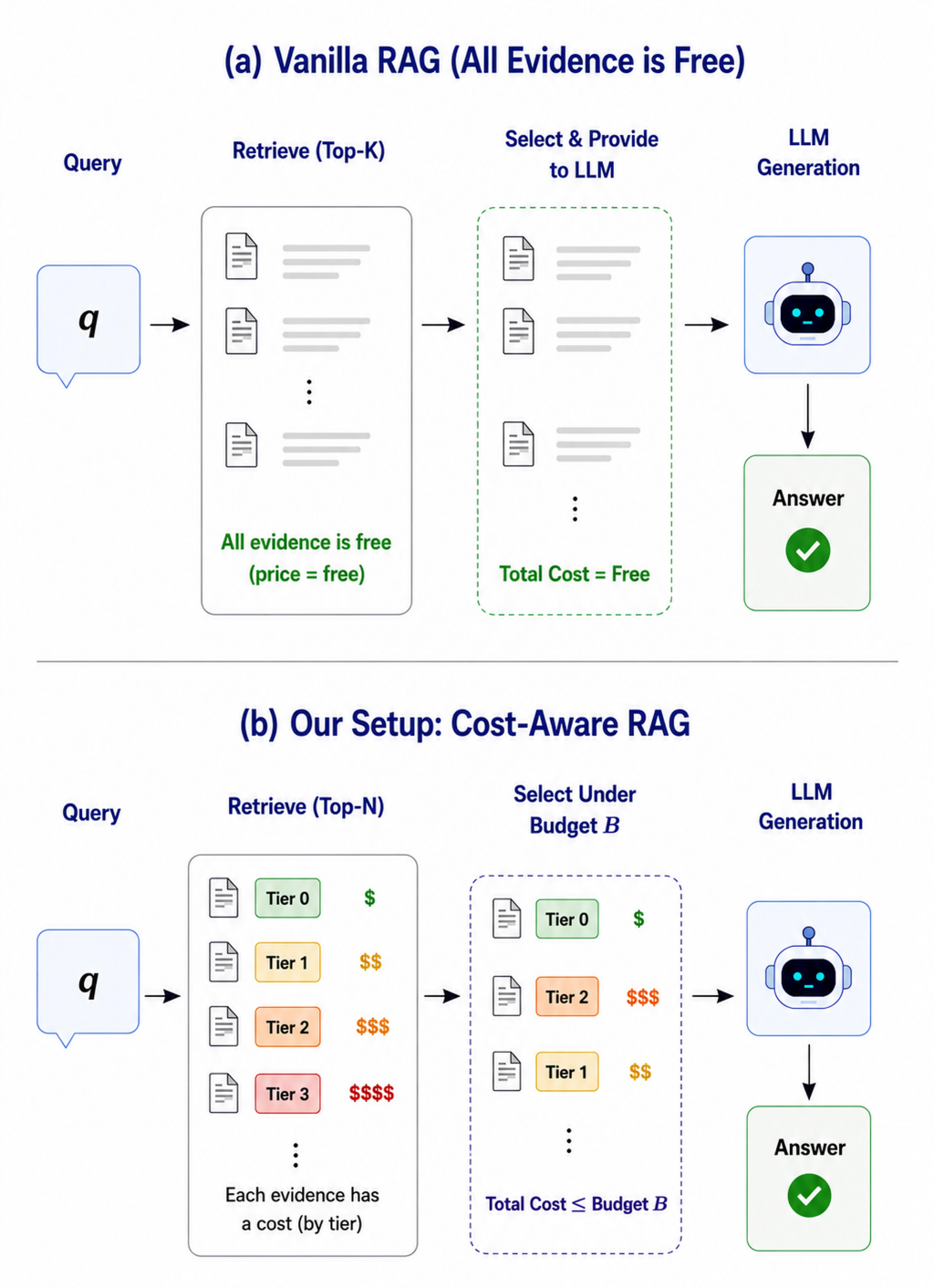}
    }
    \caption{
Comparison between vanilla and cost-aware RAG. 
(a) Vanilla RAG assumes that all retrieved evidence is freely accessible. 
(b) Our setup assigns each evidence item an access-cost tier and charges cost only for evidence selected and provided to the generator under budget \(B\).
    }
    \label{fig:cost-aware-rag-setup}
\end{figure}

%% file: latex/2_data_construction.tex
\section{Constructing a Cost-Aware Retrieval Corpus}
\label{sec:data}
\input{table/new_tier_table}

\subsection{Cost-annotated corpus.}
We build on the MS MARCO v2.1 document collection \cite{bajaj2016ms}, which consists of web-sourced passages with associated queries and source URLs. 
We keep the original queries unchanged, and augment each evidence item with an access-cost label inferred from its originating domain. These costs are intended to model relative access friction rather than literal monetary prices. Specifically, each source domain is assigned to a discrete cost tier, summarized in Table~\ref{table:access_costs_new}, ranging from open community resources to general open-web content and higher-friction curated, professional, or restricted sources. This construction enables budget-constrained retrieval and evaluation while preserving the original MS MARCO evidence and supervision.

\subsection{Domain distribution analysis.}
We first analyze the source-domain distribution of the MS MARCO v2.1 segment corpus. The corpus contains 113,520,750 segments from 1,035,806 unique domains, and the distribution is highly skewed. The most frequent domain, \texttt{en.wikipedia.org}, contributes 7,006,110 segments, corresponding to 6.17\% of the corpus, while the top 30 domains together account for 19,178,928 segments, or 16.89\% of all segments. This skew motivates a domain-level annotation strategy: a relatively small number of high-coverage domains accounts for a substantial fraction of the corpus, while most domains occur only in the long tail.

\subsection{LLM-based domain classification.}
Since manually labeling more than one million domains is impractical, we annotate the highest-coverage domains using an LLM-as-judge procedure. We select the top 20,000 domains by segment count, which together cover 82,604,646 segments, or 72.77\% of the full corpus. Given the tier definitions in Table~\ref{table:access_costs_new}, GPT-4.1-mini assigns each domain to an access-friction tier based on the domain identity and its likely accessibility characteristics. To avoid forcing arbitrary assignments, the model is also allowed to abstain when the domain cannot be confidently classified; abstentions are rare in practice (around 0.3\%). The initial annotation used a four-way rubric separating curated or professional sources from high-friction restricted sources. Under this rubric, the top 20,000 domains were classified as follows: 1,904 domains in Tier 0 (9.52\%), 14,350 in Tier 1 (71.75\%), 2,715 in Tier 2 (13.58\%), and 1,031 in Tier 3 (5.16\%). Because the two highest-friction classes are relatively sparse, we merge the original Tiers 2 and 3 into the final Tier 2 used in our experiments. This yields a three-tier structure that preserves the distinction between free, low-cost, and costly evidence while making the highest-cost tier sufficiently populated for evaluation. 

\subsection{Long-tail imputation and final tier distribution.}
The remaining domains form a large long tail with low individual coverage. Rather than querying the LLM for every rare domain, we impute their tiers by sampling from the empirical tier proportions observed among the annotated top 20,000 domains. After propagating domain labels to all segments and merging the two highest-friction tiers, the final MS MARCO-only corpus contains 36,547,305 Tier 0 segments (32.19\%), 62,632,643 Tier 1 segments (55.17\%), and 14,340,802 Tier 2 segments (12.64\%), as shown in Table~\ref{tab:msmarco_tier_distribution}. In all budgeted retrieval experiments, these tiers are mapped to costs of 0, 1, and 4, respectively.

\begin{table}[t]
\centering
\begin{tabular}{crr}
\toprule
Tier & \# Segments & Proportion \\
\midrule
0 & 36,547,305 & 32.19\% \\
1 & 62,632,643 & 55.17\% \\
2 & 14,340,802 & 12.64\% \\
\bottomrule
\end{tabular}
\caption{Final tier distribution for the cost-annotated MS MARCO v2.1 segment corpus after long-tail imputation and merging the two highest-friction tiers.}
\label{tab:msmarco_tier_distribution}
\end{table}

%% file: table/new_tier_table.tex
\begin{table}[t]
\centering
\small
\begin{tabularx}{\linewidth}{c X}
\hline
\textbf{Tier} & \textbf{Access Level} \\
\hline
0  & Open community resources (e.g., Wikipedia, public forums) \\
1  & General open-web content (blogs, tutorials, news sites) \\
2 & Curated/official/professional sources with moderate friction (official vendor docs \& API refs, support KBs, standards/gov technical docs; may include free-login or soft gating) \\
3 & High-friction restricted sources (paid paywalls, enterprise SSO portals, proprietary aggregators) \\
\hline
\end{tabularx}
\caption{Evidence cost tiers modeling relative access friction.}
\label{table:access_costs_new}
\end{table}


%% file: latex/3_methodology.tex
\section{Experimental Setting}

\subsection{Models}
We evaluate two open-weight language models: Llama-3.1-8B-Instruct~\citep{touvron2023llama} and Qwen3-8B~\citep{yang2025qwen3}. 
This allows us to compare cost-aware retrieval behavior across two widely used model families at a similar parameter scale.
During retrieval, Qwen3-embedding-0.6B~\cite{qwen3embedding} is used to retrieve query-related documents, and the number of retrieval documents is limited to the top 30.

\subsection{Datasets}
\paragraph{Retrieval corpus}
We use the cost-annotated MS MARCO v2.1 corpus as our main retrieval source~\citep{bajaj2016ms}. Each passage inherits an access-cost label from its source domain, capturing relative access friction rather than literal monetary price. In our experiments, the final cost tiers are mapped to costs $0$, $1$, and $4$, corresponding to free, low-cost, and costly evidence, respectively.

\paragraph{Evaluation benchmarks}
We evaluate on five question-answering benchmarks that span both open-domain and domain-specific settings. For open-domain QA, we use HotpotQA~\citep{yang2018hotpotqa}, Natural Questions (NQ)~\citep{kwiatkowski-etal-2019-natural}, and TriviaQA~\citep{joshi-etal-2017-triviaqa}. These benchmarks require broad web knowledge and cover different retrieval demands, from relatively direct factoid questions to multi-hop reasoning over multiple pieces of evidence. For domain-specific QA, we use MedQA-US~\citep{jin2021disease} and MMLU-Med~\citep{xiong2024acl}.

To create a more realistic high-friction retrieval scenario for the medical benchmarks, we augment the Tier-2 evidence pool for MedQA-US and MMLU-Med with Textbooks~\citep{jin2021disease}, the large medical textbook corpus released with MedQA~\citep{jin2021disease}.  We treat this corpus as costly domain-specific evidence, simulating expert medical sources such as scientific articles, textbooks, and clinical reference materials that are more expensive to access but may provide higher-quality information. This setup allows us to test cost-aware evidence selection both when relevant knowledge is broadly available on the web and when it is concentrated in specialized, high-friction sources.
We provide further experimental details in Appendix~\ref{appendix:add_exp}.

\subsection{Budgeted Retrieval Protocol}
For each query $q$, we retrieve a candidate set of passages and select a subset to provide to the generator under an evidence-access budget $B$. Each candidate passage $d_i$ is associated with a relevance score $v_i$, estimated by the similarity score between the passage and the query, and an access cost $c_i$, determined by its source cost tier. Unless otherwise stated, the generator receives at most 30 pieces of evidence per question. When the budget is exhausted before reaching this evidence limit, the remaining context slots can be filled with the highest-ranked Tier-0 evidence, since these passages incur no additional access cost. We evaluate answer quality as a function of the available budget, measuring how different selection strategies trade off relevance, cost, and redundancy. 

\subsection{Baselines and Evidence Selection Methods}
\label{sec:selection}
\paragraph{Baselines}
We compare cost-aware selection against two standard baselines. The \textbf{vanilla LLM} baseline answers without retrieved evidence. The \textbf{top-$k$} baseline provides the generator with the top-$k$ retrieved passages, ignoring access costs. These baselines separate the effect of retrieval itself from the effect of budgeted evidence selection.

\paragraph{Budgeted selectors}
We evaluate five budgeted evidence-selection methods:
\begin{itemize}
    \item \textbf{Relevance-only} ranks passages by relevance and selects the highest-scoring passages that fit within the budget.
    \item \textbf{Greedy cost-aware} ranks passages by cost-normalized relevance, prioritizing high predicted utility per unit cost.
    \item \textbf{Knapsack} formulates selection as a 0/1 knapsack problem:
    \begin{equation}
        \max_{\mathbf{x}} \sum_i v_i x_i
        \quad \text{s.t.} \quad
        \sum_i c_i x_i \leq B,
    \end{equation}
    where $x_i \in \{0,1\}$ indicates whether passage $d_i$ is selected. This directly maximizes predicted relevance under the budget constraint, while assuming independent passage utilities.
    \item \textbf{Redundancy-aware knapsack} extends the knapsack objective with a pairwise redundancy penalty:
    \begin{equation}
    \small
        \max_{\mathbf{x}} \sum_i v_i x_i-
        \eta \sum_{i<j} s_{ij} x_i x_j
        \quad \text{s.t.} \quad
        \sum_i c_i x_i \leq B,
    \label{eq:redudancy_aware_knapsack}
    \end{equation}
    where $s_{ij}$ denotes the similarity between passages and $\eta$ controls the strength of the redundancy penalty. This encourages the selector to spend budget on complementary rather than near-duplicate evidence. The $\eta$ is 0.2.
    \item \textbf{Maximum Marginal Relevance (MMR) with cost penalty} performs sequential set-aware selection. Given the selected set $S$, each remaining passage is scored as
    \begin{equation}
    \small
        m_i(S)=
        \alpha r(q,d_i)
        -
        (1-\alpha)\max_{d_j\in S} s(d_i,d_j)
        -
        \gamma c_i ,
    \label{eq:mmr}
    \end{equation}
    where $r(q,d_i)$ is the query--passage relevance score, $s(d_i,d_j)$ is passage similarity, and $\gamma$ controls the cost penalty. $\alpha$ is 0.7 and $\gamma$ is 0.1. 
    This method balances relevance, diversity, and access cost at each selection step.
\end{itemize}

\input{figure_tex/overall_figure_all}

\input{figure_tex/tier_distribution}

%% file: figure_tex/overall_figure_all.tex
\begin{figure*}[t]
    \centering
    \begin{subfigure}[b]{1.0\textwidth}
        \centering
        \includegraphics[width=\textwidth]{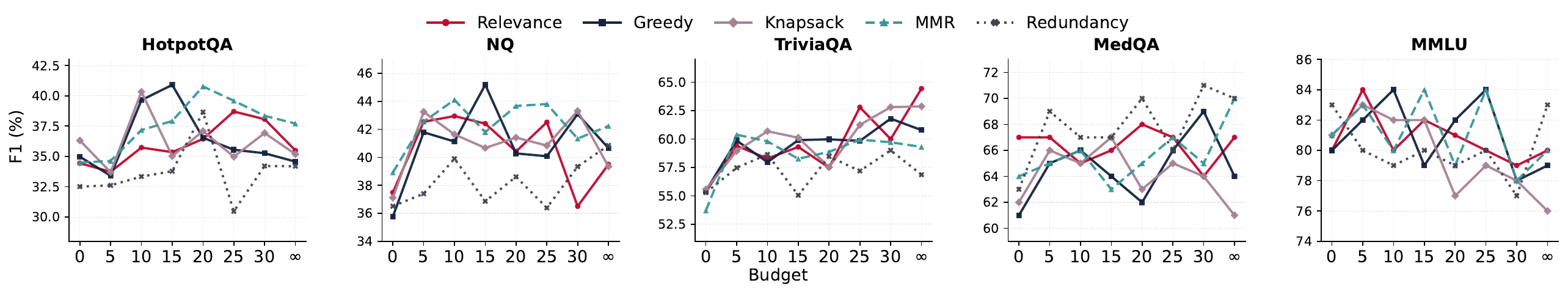}
        \caption{Performance on Qwen3-8B.}
        \label{fig:overall}
    \end{subfigure}
    
    \begin{subfigure}[b]{1.0\textwidth}
        \centering
        \includegraphics[width=\textwidth]{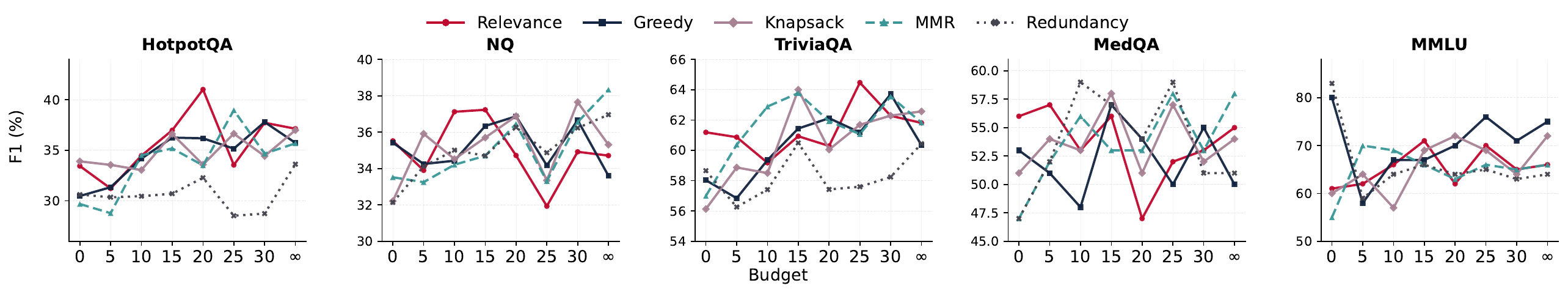}
        \caption{Performance on Llama3.1-8B-instruct.}
        \label{fig:overall_llama}
    \end{subfigure}

    \caption{Overall Performance. For more detailed results, please refer to Appendix~\ref{app:furtuer_results}.}
    \label{fig:overall_performance}
\end{figure*}

%% file: figure_tex/tier_distribution.tex
\begin{figure*}[t] 
\centering
    \includegraphics[width=0.8\textwidth]{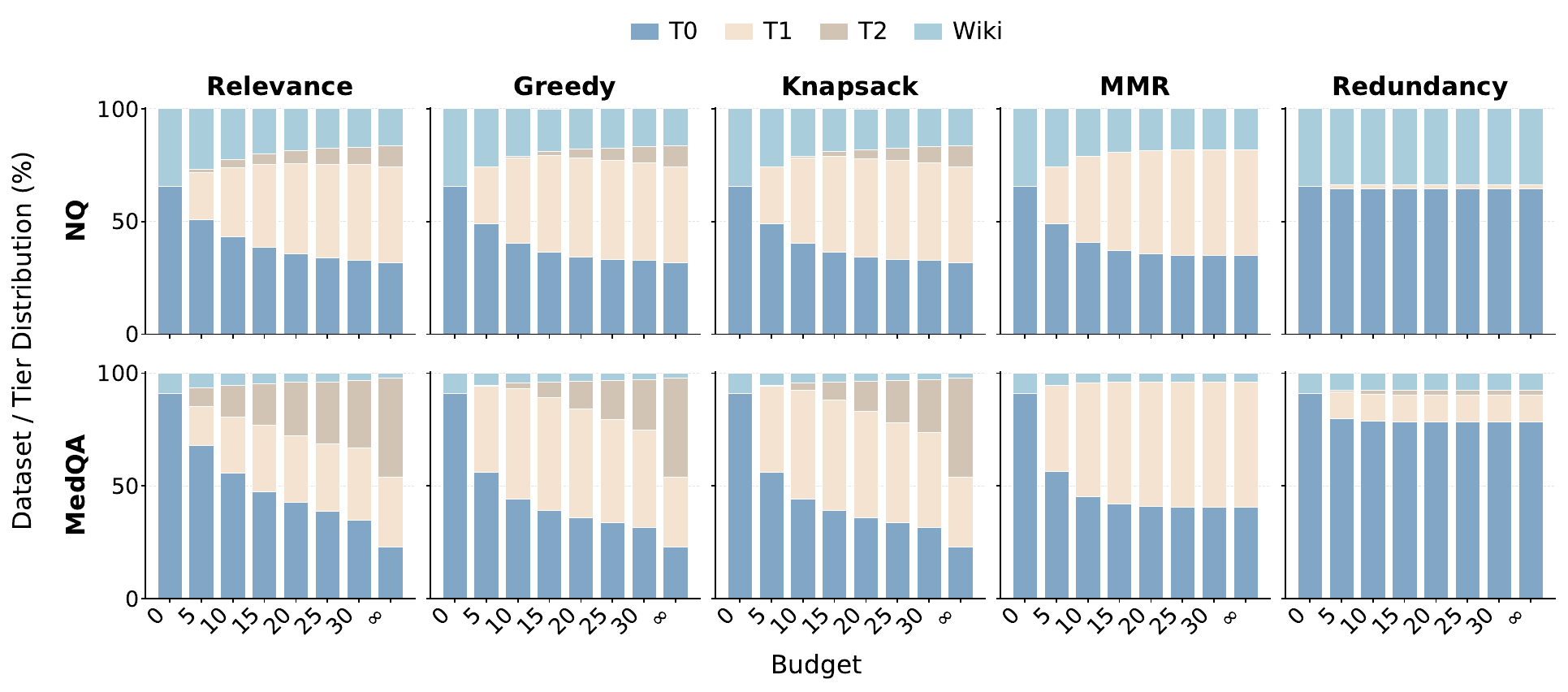}
    \caption{Tier Distribution. Wiki belongs to T0. We conduct analysis experiments using the Qwen3-8B model. The observation in Llama3.1-8B-instruct is the same, since the passages are selected by rule-based selectors. For more detailed results, please refer to Appendix~\ref{app:furtuer_results}.} \label{fig:tier_distribution}
\end{figure*}

%% file: latex/4_results.tex
\section{Results}
\label{sec:results}

We first ask whether cost-aware evidence selection can be solved by a fixed budgeted selector. The results suggest not: across datasets, budgets, and backbone models, no single rule consistently dominates, and increasing the evidence-access budget often fails to improve performance. These findings motivate a more adaptive formulation, where the system decides how much to spend and which evidence tier to access on a per-query basis.

\paragraph{No fixed selector uniformly dominates.}
Figure~\ref{fig:overall_performance} shows that the best selector depends strongly on the dataset, budget, and backbone model. With Qwen3-8B, Greedy is best on HotpotQA and NQ, reaching 40.90 and 45.19 F1 at \(B=15\), while Knapsack is best on TriviaQA and Redundancy on MedQA. On MMLU, several methods reach the best score of 84.00 at different budgets. The pattern changes with Llama3.1-8B-Instruct: Relevance is best on HotpotQA and TriviaQA, Knapsack on NQ, and Redundancy on MedQA. Thus, the relative ranking of selectors is unstable: the same objective can be competitive in one setting and suboptimal in another.

\paragraph{More budget is not necessarily better.}
Since the generator receives a fixed evidence window, increasing \(B\) mainly changes the composition of the selected evidence rather than the amount of context. The relevant question is therefore whether a more expensive evidence mix improves answer quality. It does not reliably do so. Across the five selectors and five datasets, none of the 25 Qwen3-8B curves is monotonically non-decreasing over budgets \(\{0,5,10,15,20,25,30\}\), and the same holds for Llama3.1-8B-Instruct. Moreover, only 6 out of 25 Qwen3-8B curves and 4 out of 25 Llama3.1-8B-Instruct curves achieve their best finite-budget score at \(B=30\). This remains true even in the domain-specific benchmarks, where the high-cost tier includes domain-matched medical evidence. With Qwen3-8B on MMLU-Med, Relevance reaches 84.00 F1 at \(B=5\) but drops to 79.00 at \(B=30\), while Greedy reaches 84.00 at \(B=10\) and \(B=25\) but drops to 78.00 at \(B=30\). On MedQA, Relevance and Knapsack also peak before the largest budget. Thus, even when costly evidence is plausibly useful, simply allowing more of it into the context is not enough.

\paragraph{Budget changes evidence composition differently across domains.}
Figure~\ref{fig:tier_distribution} helps explain the non-monotonic trends. In the open-domain setting, represented by NQ, larger budgets do not necessarily lead to substantially greater use of the highest-cost tier; many methods continue to rely on lower-cost evidence. This suggests that general-knowledge questions can often be answered from open or low-cost sources. In contrast, for MedQA, where the costly tier contains domain-specialized medical evidence, higher budgets often lead selectors to use more high-cost evidence. The value of expensive evidence is therefore task-dependent: conservative policies may miss useful specialized evidence, while aggressive policies may waste budget when low-cost evidence is sufficient.

\paragraph{Reducing the number of evidence passages is not sufficient.}
As shown in Table~\ref{tab:overall_performance_qwen_ci} and~\ref{tab:overall_performance_llama_ci},
the cost-agnostic top-\(k\) baseline controls for context size by providing only the \(k\) highest-ranked passages, without considering access cost. If the weakness of fixed selectors were simply that they include too many passages, then some fixed \(k\) should perform consistently well. The results do not support this explanation. Top-\(k\) sometimes benefits from using fewer than 30 passages, but no value of \(k\) is consistently strong across datasets or models. With Qwen3-8B, the best top-\(k\) scores on HotpotQA and NQ are 37.67 and 40.09 F1, below the best budgeted-selector scores of 40.90 and 45.19 F1. With Llama3.1-8B-Instruct, top-\(k\) also falls short on HotpotQA and NQ. Thus, shorter contexts alone do not explain the gains; the harder problem is deciding which evidence is worth including and from which cost tier.

\paragraph{Implication: cost-aware RAG requires adaptive allocation.}
Taken together, these results show that cost-aware evidence selection is difficult to reduce to a single fixed rule. The best selector varies across datasets, models, and budgets; larger budgets do not reliably help; and the usefulness of costly evidence depends on the domain and query. This motivates a sequential formulation in which the system assesses the evidence already acquired, decides whether it is sufficient, and determines whether additional or higher-cost evidence is worth the expense. We therefore turn to an agentic formulation of cost-aware RAG, where evidence acquisition is treated as an online decision process: the model must decide not only what to retrieve, but also when to spend, where to spend, and when to stop.

%% file: latex/5_agent.tex
\input{figure_tex/agent}
\section{Agentic Cost-Aware Evidence Acquisition}
\label{sec:agent}

The results in Section~\ref{sec:results} suggest that fixed budgeted selectors are too brittle for cost-aware RAG. We therefore study an LLM-based retrieval agent as a testbed for adaptive evidence acquisition. At inference time, the agent must decide whether to retrieve, which cost tier to access, and when the accumulated evidence is sufficient to answer. This setting is more flexible than static selection, but it also introduces new failure modes: the agent may stop prematurely, spend too aggressively, or retrieve from an inappropriate tier. We therefore view agentic RAG not as a complete solution, but as a probe of whether current LLMs can act as cost-sensitive evidence controllers.

\input{figure_tex/agent_overall_allinone}
\input{figure_tex/agent_budget_cost_f1_fig}
\paragraph{Design.}
We instantiate the agent with a ReAct-style loop \cite{DBLP:conf/iclr/YaoZYDSN023}. At each step, the agent observes the question, collected evidence, remaining budget, and available tiers with their costs. It then chooses between two actions: \textsc{Retrieve}, which specifies a tier and returns one passage from that tier ranked by query similarity, or \textsc{Answer}, which terminates the loop (see Figure~\ref{fig:agentic_cost_aware_loop}). The agent uses the same corpus, retriever, and cost tiers as the fixed-selector experiments; only the acquisition policy changes. The prompt instructs the agent to obey the budget and answer only when the evidence is sufficient or no useful retrieval is affordable. The full prompt is shown in Table~\ref{tab:agent_prompt_all}. Unless otherwise stated, we use a per-query budget of \(B=20\) with the same two backbones as before, Qwen3-8B and Llama3.1-8B-Instruct. The budget is an upper bound; we report the average realized cost. We also evaluate a \emph{shared-budget} variant in which the agent receives batches of five questions and allocates a total budget of 100 before answering. This diagnostic setting tests whether the model can identify, at the workload level, which questions are likely to require costly evidence. Additional details are provided in Appendix~\ref{appendix:add_exp}.

\paragraph{Per-query agents can be efficient evidence controllers.}
Figure~\ref{fig:agent_result_allinonw} compares per-query agents with fixed budgeted selectors. The main result is that agentic acquisition can often match or exceed fixed selectors while using far fewer passages and much lower realized cost. This improvement is not explained by shorter contexts alone: as shown in Section~\ref{sec:results}, cost-agnostic top-\(k\) selection does not consistently perform well. The clearest gains come from Qwen3-8B. On NQ, the agent reaches 49.93 F1 with only 0.50 average cost and 2.47 retrieved passages, outperforming all fixed selectors while spending far less than the high-budget settings. On TriviaQA, it reaches 66.73 F1 with 0.52 cost and 2.60 passages, and on MMLU-Med it reaches 85.00 F1 with 2.90 cost and only 1.14 passages. On MedQA, it obtains 68.00 F1 with 5.08 cost and 1.28 passages, competitive with the best fixed selectors despite using a much smaller evidence context. These results suggest that the agent is not simply retrieving more evidence; rather, it often stops after acquiring a small amount of useful evidence.

\paragraph{Agentic cost awareness is model- and task-dependent.}
Agentic control is not automatically cost-aware. HotpotQA remains difficult: with Qwen3-8B and \(B=20\), the agent reaches only 30.34 F1, below the best fixed selector. This suggests that multi-hop questions may require more than repeated retrieval with the original query, such as query decomposition, sub-question planning, or targeted query rewriting. \textbf{Backbone choice also matters}. Llama3.1-8B-Instruct can achieve competitive scores, but often by spending much more. With \(B=20\), it reaches 46.77 F1 on NQ using 13.47 average cost and 12.54 passages, and 68.40 F1 on TriviaQA using 11.53 cost and 11.53 passages. On MedQA, it reaches 60.00 F1 with 18.39 cost, and on MMLU-Med it reaches only 72.00 F1, substantially below the best fixed selectors. In contrast to Qwen3-8B, the Llama agent often treats the available budget as a resource to consume rather than a constraint to manage. Thus, the key challenge is not only choosing the right tier, but also estimating when the current evidence is sufficient.

\paragraph{Budget sensitivity reveals evidence-sufficiency failures.}
\label{sec:agent_result_budget}

A cost-aware agent should not spend more merely because more budget is available. We therefore vary the per-query budget on NQ and MedQA, covering both open-domain and domain-specific settings. Figure~\ref{fig:agent_budget_cost_f1} shows that increasing the budget has limited effect on final F1, but substantially changes spending behaviour. Qwen3-8B exhibits relatively stable stopping behaviour. Increasing the budget from \(B=20\) to \(B=30\) changes its average cost from 0.50 to 1.06 on NQ and from 5.08 to 4.72 on MedQA, while F1 changes from 49.93 to 45.20 and from 68.00 to 69.00. In contrast, Llama3.1-8B-Instruct spends much more as the budget increases, without corresponding gains: from \(B=20\) to \(B=30\), its cost rises from 13.47 to 15.54 on NQ and from 18.39 to 27.01 on MedQA, while F1 drops from 46.77 to 42.89 and from 60.00 to 56.00. These trends indicate that evidence-sufficiency estimation is a central bottleneck for agentic cost-aware RAG. Representative trajectories are provided in Appendix~\ref{app:agent_result}
and a corresponding case study comparing this behaviour across both models is presented in Appendix~\ref{app:agent_casestudy}.

\paragraph{Shared-budget allocation probes workload-level planning.}
\label{sec:agent_result_batch}

We next evaluate shared-budget allocation as a diagnostic for workload-level
planning. Unlike the per-query setting, the agent receives a batch of questions
and must decide in advance which ones deserve more budget. For Qwen3-8B, this
improves the performance--cost tradeoff on most datasets: compared with the
per-query \(B=20\) agent, it reduces cost and passage usage on HotpotQA,
TriviaQA, MedQA, and MMLU-Med while improving or maintaining F1. For example,
TriviaQA improves from 66.73 to 70.27 F1 while cost drops from 0.52 to 0.33;
MedQA improves from 68.00 to 71.00 F1 while cost drops from 5.08 to 4.38; and
MMLU-Med maintains 85.00 F1 while reducing cost from 2.90 to 2.56. However,
the benefit is not uniform: on NQ, Qwen spends more while performing worse, and for Llama3.1-8B-Instruct, lower cost and passage usage often come with lower
F1. Thus, shared-budget allocation can help when the model estimates question
difficulty well, but it also introduces a new failure mode: the agent must
predict before retrieval which questions are worth spending on.

\paragraph{Takeaway.}
Agentic frameworks naturally fit cost-aware RAG: each retrieval decision can be
conditioned on the question, the evidence collected so far, and the remaining
budget. This adaptivity can improve the performance--cost tradeoff when the
model recognizes that a small amount of evidence is sufficient. However,
current agents remain unreliable cost-sensitive controllers: their behaviour
varies with the backbone, task structure, and evidence-sufficiency judgments.
Cost-aware RAG therefore highlights a capability underexplored in standard RAG:
deciding whether additional evidence is worth its access cost.

%% file: figure_tex/agent.tex
\begin{figure}[t]
    \centering
    \includegraphics[width=0.7\columnwidth]{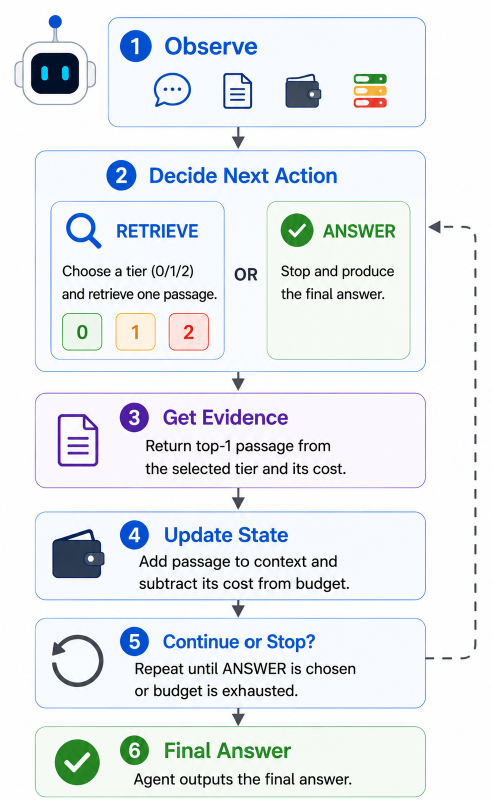}
    \caption{Agentic cost-aware evidence acquisition loop.}
    \label{fig:agentic_cost_aware_loop}
\end{figure}

%% file: figure_tex/agent_overall_allinone.tex
\begin{figure*}[t] 
\centering
    \includegraphics[width=0.8\textwidth]{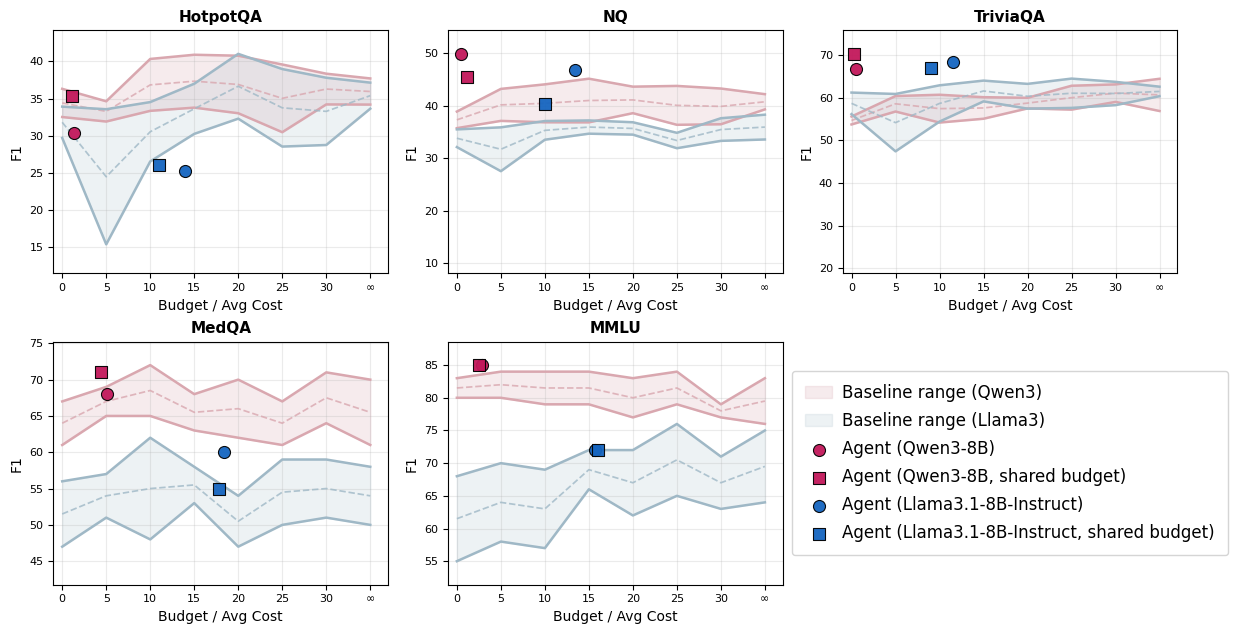}
    \caption{Agent Overall Performance. We compare agent performance with evidence selection method baselines.} \label{fig:agent_result_allinonw}
\end{figure*}

%% file: figure_tex/agent_budget_cost_f1_fig.tex

\begin{figure}[t]
    \centering

    \begin{subfigure}[t]{0.49\columnwidth}
        \centering
        \includegraphics[width=\linewidth]{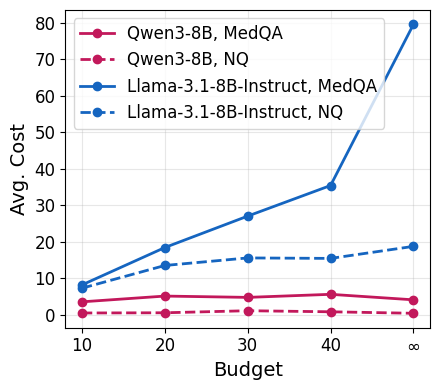}
        \caption{Average Cost}
        \label{fig:agent_budget_cost}
    \end{subfigure}
    \begin{subfigure}[t]{0.49\columnwidth}
        \centering
        \includegraphics[width=\linewidth]{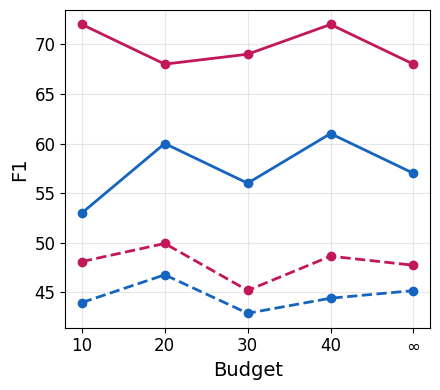}
        \caption{F1}
        \label{fig:agent_budget_f1}
    \end{subfigure}

    \caption{Cost and F1 with Increasing Budget for Agent on NQ and MedQA.}
    \label{fig:agent_budget_cost_f1}
\end{figure}

%% file: latex/7_related_works.tex
\section{Related Work}
\label{sec:related_work}

\paragraph{RAG under resource constraints.}
Most RAG work assumes that retrieved evidence is freely accessible
\cite{lewis2020retrieval,guu2020retrieval,izacard2021leveraging}. Recent work
studies resource-constrained RAG by limiting context length, token usage,
latency, inference-time computation, or the number of retrieved chunks
\cite{jeong2024adaptive,yue2025inference,lu-etal-2025-turborag}. Related
budgeted-RAG work attaches cost to later stages of the pipeline: CORAG selects
among already-retrieved chunks under a cost constraint \cite{wang2024corag},
while Chunk-as-a-Service studies pricing for chunks used to enrich prompts
\cite{al2026budget}. Our setting moves the cost earlier: evidence may be
costly to access depending on its source, so the system must decide whether a
higher-cost source is worth acquiring before its evidence can be used.

\paragraph{Adaptive and agentic RAG.}
Adaptive and agentic RAG methods make retrieval conditional rather than fixed:
Adaptive-RAG routes queries by estimated complexity
\cite{jeong2024adaptive}, Active-RAG/FLARE and Self-RAG trigger retrieval
during generation \cite{jiang2023active,asai2024self}, and ReAct interleaves
reasoning with actions and observations \cite{DBLP:conf/iclr/YaoZYDSN023}.
These methods motivate our agentic formulation, but they do not study evidence
access costs. In our setting, the model must decide not only whether more
evidence is needed, but whether it is worth acquiring from a low- or high-cost
source.

%% file: latex/5_conclusion.tex
\section{Conclusion}
\label{sec:conclusion}

We introduced cost-aware RAG, where evidence carries source-dependent access
costs and retrieval becomes a budgeted acquisition problem. Across open-domain
and domain-specific QA, fixed selectors are brittle: no rule consistently
dominates, and higher budgets do not reliably improve answer quality. Agentic
RAG offers a promising alternative by deciding when to retrieve, which tier to
access, and when to stop, but current agents remain highly model- and
task-dependent. 

%% file: latex/6_limitation.tex
\section{Limitations}

\paragraph{Simulated access costs.}
Our access-cost tiers are proxies for relative access friction rather than direct measurements of monetary cost, licensing restrictions, latency, or institutional access. We assign costs at the domain level and map them to a small number of discrete tiers, which enables large-scale evaluation but cannot capture variation within the same source.

\paragraph{Automatic tier annotation.}
Our cost annotations rely on automatic domain classification and long-tail imputation. The LLM judge may misclassify some domains, and imputed labels for rare domains may not reflect their true access properties. These errors could affect the apparent value of high-cost evidence and the behavior of cost-aware selectors.

\paragraph{Evaluation scope.}
Our evaluation is limited to five QA benchmarks and sampled subsets. This allows controlled comparisons across selectors, budgets, and backbones, but leaves open how stable the conclusions are on larger samples, additional domains, and different answer formats. The medical benchmarks should be viewed as domain-specific QA testbeds rather than evidence of clinical readiness.

\paragraph{Simplified retrieval environment.}
The agent retrieves from a fixed candidate pool using the original question and receives one passage at a time from a selected tier. It does not perform query rewriting, sub-question decomposition, source verification, or multi-query search, which likely limits performance on multi-hop questions.

\paragraph{Zero-shot agent control.}
Our agentic controller is evaluated in a zero-shot prompting regime with two open-weight backbones. The results show that agentic cost-aware RAG can be efficient, but stopping and spending behavior remains highly model-dependent. We therefore view the agentic results as a diagnostic of current LLMs as cost-sensitive evidence controllers, rather than as a fully optimized policy.

%% file: latex/8_appendix.tex
\appendix

\section{License}
This section summarizes the licenses of the datasets used in our experiments.

All datasets used in this work permit academic use under their respective licenses and agreements: Natural Questions (CC-BY-SA-3.0 License); HotpotQA (CC-BY-4.0 License); TriviaQA (Apache 2.0 License); MedQA-US and MMLU-Med are under the terms of the United States Copyright Act.

\input{table/agent_result_all_table_ci}

\section{Additional Experimental Details}
\label{appendix:add_exp}

\paragraph{Implementation Details.}
We conducted our experiments on an NVIDIA H100 NVL GPU with 94 GB of memory. We used vLLM~\cite{vllm} as the inference engine to accelerate the inference process. 

\label{app:add_exp_data}

\paragraph{Corpus Dataset.}
We use the MS~MARCO~v2.1\footnote{\url{https://trec-rag.github.io/annoucements/2024-corpus-finalization/}} corpus as our main retrieval source~\citep{bajaj2016ms} for general QA tasks. 
For domain-specific tasks we use Textbooks\footnote{\url{https://github.com/jind11/MedQA}}~\cite{jin2021disease} as knowledge base. Textbooks is a collection of 18 widely used English medical books, which were originally collected in PDF format and converted into digital text via OCR followed by clean-up processing. The dataset contains 231,581 paragraphs with an average paragraph length of 55.0 and a maximum paragraph length of 1,234. 
We select this corpus for our domain-specific tasks because it has been shown to effectively support MedQA and MMLU-Med, according to \citep{xiong2024acl}. We access the Textbooks corpora in version of MIRAGE\footnote{\url{https://github.com/gzxiong/MIRAGE}}~\cite{xiong2024acl}. 






\paragraph{Evaluation Benchmarks \& Metrics.}
We evaluate on five question-answering benchmarks that span both open-domain and domain-specific settings. 
For open-domain QA, we use HotpotQA~\citep{yang2018hotpotqa}, Natural Questions~\citep{kwiatkowski-etal-2019-natural}, and TriviaQA~\citep{joshi-etal-2017-triviaqa}. 
For domain-specific QA, we use MedQA-US~\citep{jin2021disease} and MMLU-Med~\citep{xiong2024acl}.

We have access to general-domain tasks via the KILT Benchmarking
\footnote{\url{https://trec-rag.github.io/annoucements/2024-corpus-finalization/}}~\cite{KILT},
and domain-specific tasks from their individual GitHub repositories.
Following \citet{DBLP:conf/emnlp/WangZYZCYLWWHLS25} and considering our computational resources and the difficulty of conducting case-by-case investigation, we randomly sample 100 questions from each benchmark. 
We evaluate performance using Token-Level F1 (F1) and Exact Match (EM). 
While the F1 score measures the trade-off between coverage and redundancy by allowing partial matches, EM imposes a stricter criterion, requiring the model's output to perfectly match the reference answer. 
If the benchmark provides multiple reference answers, we adopt the highest score. 
For each benchmark and method, token-level F1 is first computed for each question and then averaged over the 100 sampled questions. We compute the 95\% confidence interval for the mean F1 using a normal approximation, where the standard error is estimated from the sample variance of question-level F1 scores.
Values in the tables are reported as the mean F1 $\pm$ the corresponding 95\% confidence-interval margin.

\input{figure_tex/agent_budget_psg_em_fig}

\paragraph{Hyperparameter Search for Budgeted Selectors.}
\label{app:add_exp_para}
\input{figure_tex/app_para_fig}

We present the hyperparameter selection for the Budget Selectors discussed in Section~\ref{sec:selection}. 
In Equation~\ref{eq:redudancy_aware_knapsack}, $\eta$ controls the strength of the redundancy penalty in the Redundancy-aware Knapsack method. 
The hyperparameters $\alpha$ and $\gamma$ in Equation~\ref{eq:mmr} balance query--passage relevance against passage similarity and control the cost penalty, respectively. 
Figure~\ref{fig:parameter} reports the F1 scores of these methods under different hyperparameter settings on the NQ dataset. The results show that the best overall performance on both benchmarks is achieved with $\eta=0.2$ for Redundancy-aware Knapsack, and with $\alpha=0.7$ and $\gamma=0.1$ for MMR.

\paragraph{Prompt Templates in Agentic Framework.}
\label{app:add_exp_agent}
\input{table/agent_prompt_all_table}

\input{table/agent_prompt_budget_table}

We provide the full prompts used for the agent in Table~\ref{tab:agent_prompt_all} and Table~\ref{tab:agent_prompt_budget}. Specifically, Table~\ref{tab:agent_prompt_all} presents the prompt for the agent with a per-query budget, while Table~\ref{tab:agent_prompt_budget} presents the prompt for budget allocation in a shared budget setting.

\paragraph{Tier Annotation Consistency with Human.} We manually inspected 30 randomly sampled tier assignments produced by GPT-4.1-mini. Most labels were considered reasonable, and disagreements mainly occurred between adjacent tiers, suggesting that the generated tier annotations are generally consistent for our analysis purposes.

\section{Additional Experimental Results}
\label{app:furtuer_results}

\paragraph{Full Result of Evidence Selection Methods.}
Figure~\ref{fig:overall_performance} shows different evidence selection methods across datasets and backbone models. We provide more detailed overall performance Tables in Table~\ref{tab:overall_performance_qwen_ci} and~\ref{tab:overall_performance_llama_ci}, containing Vanilla LLM and Topk RAG performance methods. In addition, Figure~\ref{fig:tier_distribution} presents the tier distribution of retrieved documents on NQ and MedQA datasets. We provide the full result of tier distribution across all datasets in Figure~\ref{fig:full_tier_distribution}.

\paragraph{Full Result of Agentic framework.}
\label{app:agent_result}
Figure~\ref{fig:agent_result_allinonw} presents the agent performance compared with evidence selection baselines. We report more detailed results of the agent in Table~\ref{tab:agentic_rag_overall}, containing Avg. Cost, Avg. Psg, and Avg. Steps.

In Section~\ref{sec:agent_result_budget}, we discuss the sensitivity of the agent to increasing budgets. Further analyze the impact on the average number of retrieved passages and the EM score in Figure~\ref{fig:agent_budget_psg_em}. We observe a similar trend, that the Llama3.1-8B-Instruct model tends to retrieve more passages with increasing budget, while the Qwen3-8B model remains stable. Also, the increasing budget does not lead to a higher EM score.

\paragraph{Additional Experiment with Alternative Pricing Setting.}
In our main experimental setting, we employ the pricing setting where tier 0 is free, tier 1 costs 1, and tier 2 costs 4. We additionally conduct experiments under an alternative pricing setting, where tier 0 is free, tier 1 costs 1, and tier 2 costs 2. As shown in Table~\ref{tab:overall_performance_qwen_costmore_ci}, the main conclusions are aligned with the results observed in Section~\ref{sec:results}.

\section{Case Studies: Agents' Cost-aware Decision Making}
\label{app:agent_casestudy}

We illustrate how the agent processes tasks. In the first case, as shown in Table~\ref{tab:agent-case-study}, 
the agent first identifies that the question requires external evidence about a specific historical election and therefore begins with tier 0, the free retrieval source. When the first retrieved passage fails to provide information about the target year,
it continues retrieving from tier 0, since the needed fact is likely available in encyclopedic sources. The second retrieval provides sufficient evidence that Eliot Cutler finished ahead of Libby Mitchell in Maine's 2010 gubernatorial election, allowing the agent to answer correctly.
This example shows that the agent performs tier-aware evidence acquisition. It keeps retrieving when evidence is insufficient, avoids unnecessarily expensive sources, and reaches the correct answer.


In the second case, as shown in Table~\ref{tab:agent-case-study-2}, the agent handles a question that requires specific and up-to-date factual evidence about Mount Ruapehu's most recent eruption.
The agent first retrieves from tier 0, since the question asks for a factual answer that may be available in free community resources. 
However, the initial passages only provide historical information, rather than the exact date of the latest eruption.
After multiple unsuccessful attempts from tier 0 and tier 1, the agent decides to explore a higher-cost tier, since cheaper sources have not yet provided the required evidence and the remaining budget is still sufficient. Although the tier 2 retrieval also fails to directly answer the question, the agent does not stop prematurely. Instead, it continues retrieval and eventually finds a useful tier 0 passage that explicitly states the last eruption date as 25 September 2007.
This case shows that the agent can adaptively balance cost and evidence quality. 
It incurs a total cost of 7 and switches across all three tiers, but eventually retrieves the necessary passage and stops retrieving once sufficient evidence is found.

In the third case, as shown in Table~\ref{tab:agent-case-study-3}, the agent handles a domain-specific medical question and correctly identifies tier 2 as the appropriate source for professional medical knowledge. Since the retrieved passages do not fully specify the protocol for this situation, the agent remains uncertain and continues searching until the budget is exhausted. With no remaining budget, the agent then combines the incomplete evidence with its own parametric knowledge and reasoning to make a correct final decision. 
This case shows that the agent can identify the appropriate knowledge tier, understand the trade-off between task difficulty and budget, and continue evidence acquisition when confidence is insufficient.


We further demonstrate how the agent performs flexible budget allocation under the shared-budget setting. As shown in Table~\ref{tab:agent-case-study-4}, the agent first estimates the domain, difficulty, and likely retrieval tier of each question. It assigns smaller budgets to questions that can likely be answered by public or low-cost sources, while reserving larger budgets for questions that require more specific historical or scientific knowledge. The agent then iteratively adjusts the allocation until the total budget exactly matches the shared budget constraint. This case shows that the agent can reason about the relative difficulty of multiple questions and distribute the available budget accordingly.

We further compare cost-aware behaviours across different backbone models. As shown in Table~\ref{tab:agent-case-study-medqa-qwen} and Table~\ref{tab:agent-case-study-medqa-llama}, both agents identify the question as requiring domain-specific medical knowledge and therefore select tier 2 for retrieval. However, they differ substantially in when to stop retrieving. The agent with Qwen3-8B retrieves only one tier 2 passage, recognizes that the evidence is sufficient, and stops with a total cost of 4. 
In contrast, the agent with Llama3.1-8B-Instruct repeatedly retrieves from tier 2 despite already having relevant evidence, and only answers after exhausting the full budget of 40. 
This comparison suggests that cost-aware behaviour is strongly influenced by the backbone model. 

\input{table/agent_casestudy_1_table}
\input{table/agent_casestudy_2_table}
\input{table/agent_casestudy_3_table}
\input{table/agent_casestudy_4_table}

\input{table/agent_casestudy_5_medqa_qwen_table}
\input{table/agent_casestudy_6_medqa_llama_table}


\input{figure_tex/tier_distribution_full}

\input{table/overall_performance_qwen_ci}
\input{table/overall_performance_llama_ci}
\input{table/overall_costmore_ci}

%% file: table/agent_result_all_table_ci.tex
\begin{table*}[t]
\centering
\small
\resizebox{\textwidth}{!}{
\begin{tabular}{ll l ccc cc}
\toprule
\textbf{Model} & \textbf{Budget Strategy} & \textbf{Dataset} 
& \textbf{Avg. Cost} & \textbf{Avg. Psg} & \textbf{Avg. Steps} 
& \textbf{EM (\%)} & \textbf{F1 (\%)} \\
\midrule

\multirow{10}{*}{Qwen3-8B}
& \multirow{5}{*}{Per-query Budget, $B=20$}
& HotpotQA & 1.38 & 6.35 & 7.41 & 20.00 & $30.34_{\pm 8.15}$ \\
& & NQ       & 0.50 & 2.47 & 3.49 & 37.00 & $49.93_{\pm 8.65}$ \\
& & TriviaQA & 0.52 & 2.60 & 3.61 & 64.00 & $66.73_{\pm 9.05}$ \\
& & MedQA    & 5.08 & 1.28 & 2.29 & 68.00 & $68.00_{\pm 9.19}$ \\
& & MMLU     & 2.90 & 1.14 & 2.14 & 85.00 & $85.00_{\pm 7.03}$ \\
\cmidrule(lr){2-8}

& \multirow{5}{*}{Shared Budget}
& HotpotQA & 1.08 & 5.66 & 6.70 & 25.00 & $35.30_{\pm 8.61}$ \\
& & NQ       & 1.14 & 2.21 & 3.21 & 33.00 & $45.52_{\pm 8.72}$ \\
& & TriviaQA & 0.33 & 2.36 & 3.37 & 66.00 & $70.27_{\pm 8.63}$ \\
& & MedQA    & 4.38 & 1.13 & 2.13 & 71.00 & $71.00_{\pm 8.94}$ \\
& & MMLU     & 2.56 & 0.86 & 1.86 & 85.00 & $85.00_{\pm 7.03}$ \\

\midrule

\multirow{10}{*}{Llama3.1-8B-Instruct}
& \multirow{5}{*}{Per-query Budget, $B=20$}
& HotpotQA & 13.97 & 13.16 & 14.56 & 15.00 & $25.21_{\pm 7.68}$ \\
& & NQ       & 13.47 & 12.54 & 13.74 & 35.00 & $46.77_{\pm 8.64}$ \\
& & TriviaQA & 11.53 & 11.53 & 12.81 & 63.00 & $68.40_{\pm 8.70}$ \\
& & MedQA    & 18.39 & 4.84  & 5.85  & 60.00 & $60.00_{\pm 9.65}$ \\
& & MMLU     & 15.74 & 7.02  & 8.11  & 72.00 & $72.00_{\pm 8.84}$ \\
\cmidrule(lr){2-8}

& \multirow{5}{*}{Shared Budget}
& HotpotQA & 10.97 & 10.12 & 11.40 & 15.00 & $25.99_{\pm 7.69}$ \\
& & NQ       & 10.00 & 9.87  & 11.04 & 28.00 & $40.41_{\pm 8.59}$ \\
& & TriviaQA & 9.02  & 8.80  & 10.06 & 63.00 & $66.87_{\pm 8.94}$ \\
& & MedQA    & 17.84 & 4.72  & 5.74  & 55.00 & $55.00_{\pm 9.80}$ \\
& & MMLU     & 16.04 & 6.28  & 7.35  & 72.00 & $72.00_{\pm 8.84}$ \\

\bottomrule
\end{tabular}}
\caption{Overall Performance of Agent across Backbone Models and Budget Strategies. For per-query budget, the budget (B) is 20; for shared budget, we set the budget as 100 per batch. }
\label{tab:agentic_rag_overall}
\end{table*}

%% file: figure_tex/agent_budget_psg_em_fig.tex
\begin{figure}[t] 
\centering
    \includegraphics[width=1.0\columnwidth]{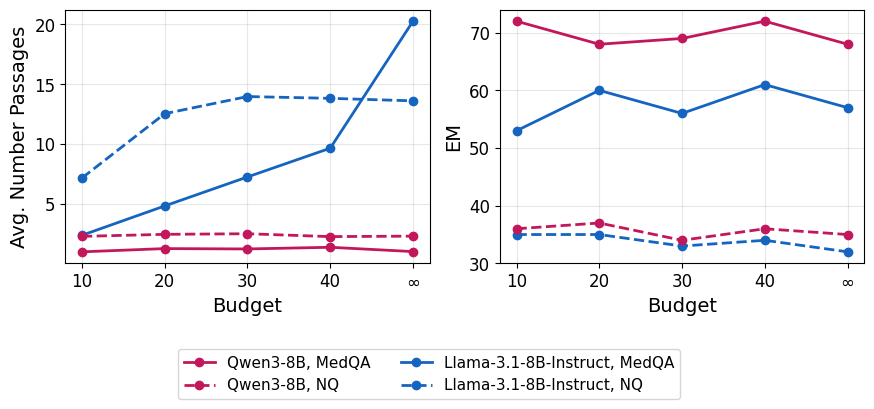}
    \caption{Number of Passages and EM with Increasing Budget for Agent on NQ and MedQA. Left: average number of passages; Right: EM score.}
    \label{fig:agent_budget_psg_em}
\end{figure}

%% file: figure_tex/app_para_fig.tex
\begin{figure*}[t]
    \centering

    \begin{subfigure}[t]{0.32\textwidth}
        \centering
        \includegraphics[width=\linewidth]{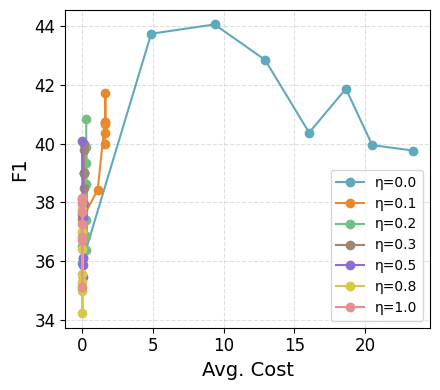}
        \caption{}
        \label{fig:para_nq_eta}
    \end{subfigure}
\begin{subfigure}[t]{0.32\textwidth}
        \centering
        \includegraphics[width=\linewidth]{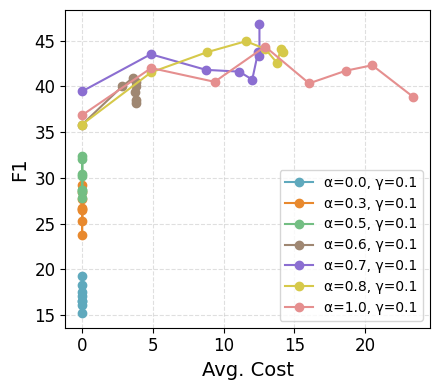}
        \caption{}
        \label{fig:nq_alpha}
    \end{subfigure}
\begin{subfigure}[t]{0.32\textwidth}
        \centering
        \includegraphics[width=\linewidth]{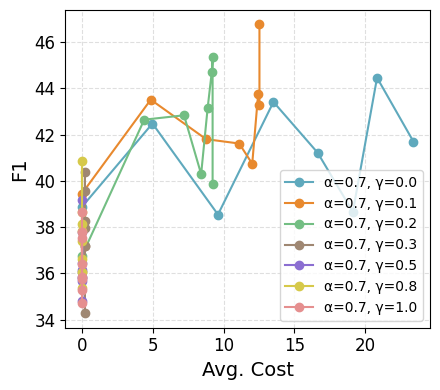}
        \caption{}
        \label{fig:nq_gamma}
    \end{subfigure}
\hspace{0.005\textwidth}
    \caption{
Hyperparameter search for Budgeted Selectors. Subfigure~\ref{fig:para_nq_eta} shows the selection of $\eta$ for the Redundancy-aware Knapsack method. Subfigure~\ref{fig:nq_alpha} shows the selection of $\alpha$ for MMR method. Subfigure~\ref{fig:nq_gamma} shows the selection of $\gamma$ for MMR method.
    }
    \label{fig:parameter}
\end{figure*}

%% file: table/agent_prompt_all_table.tex
\begin{table}[t]
\centering
\small
\begin{tabularx}{\columnwidth}{|X|}
\hline
\rowcolor[gray]{0.9} \textbf{Prompt} \\ \hline
\texttt{You are a budgeted retrieval-augmented QA agent.} \\
\texttt{At each step, first decide whether the question requires external knowledge beyond the current context.} \\
\texttt{If external knowledge is needed and budget allows, retrieve exactly ONE passage from the most appropriate tier.} \\
\texttt{Actually, you can retrieve the passages from the same tier many times.} \\
\texttt{Only answer when the current context is sufficient, the question can be answered confidently without retrieval,} \\
\texttt{or the remaining budget cannot support a useful retrieval.} \\
\texttt{Do not answer merely because the current context is empty; at the beginning, empty context is normal.} \\
\texttt{Follow the budget strictly. Output MUST be a single JSON object.} \\
\\
\texttt{Budget: \textless budget\textgreater{} \textbar{} Spent: \textless spent\textgreater{} \textbar{} Remaining: \textless remaining\textgreater{}} \\
\texttt{Question: \textless question\textgreater{}} \\
\texttt{Tier options: \textless tier\_desc\textgreater{}} \\
\texttt{Trajectory so far: \textless traj\_block\textgreater{}} \\
\texttt{Current context: \textless ctx\_block\textgreater{}} \\
\texttt{Decision objective:} \\
\texttt{- First judge whether answering requires outside knowledge.} \\
\texttt{- If outside knowledge is needed, retrieve one passage from the best tier.} \\
\texttt{- Continue retrieving across steps until the evidence is sufficient, or the budget is exhausted.} \\
\texttt{- If the question can already be answered confidently from the question itself or the accumulated context, answer directly.} \\
\texttt{Decide the next action. Choose one:} \\
\texttt{1) answer now} \\
\texttt{2) retrieve one passage from a tier (0/1/2)} \\
\texttt{Return JSON in one line with schema:} \\
{\textsc{action}: \textsc{answer} $\mid$ \textsc{retrieve}, \textsc{tier}: 0 $\mid$ 1 $\mid$ 2 $\mid$ null, \textsc{reason}: string}\\
\texttt{Rules:} \\
\texttt{- If action=answer, set tier=null.} \\
\texttt{- If action=retrieve, set tier to 0/1/2.} \\
\texttt{- If Remaining \textless{} cost(tier), you MUST answer.} \\
\texttt{- Do not answer just because Current context is empty.} \\
\texttt{- If the information is insufficient, retrieve before answering.} \\
\texttt{- Prefer the cheapest tier likely to be sufficient, but use higher tiers when the question likely needs official, professional, or restricted-domain knowledge.} \\

 \hline
\end{tabularx}
\caption{Prompt Template for Per-query Budget Agent.} 
\label{tab:agent_prompt_all}
\end{table}

%% file: table/agent_prompt_budget_table.tex
\begin{table}[t]
\centering
\small
\begin{tabularx}{\columnwidth}{|X|}
\hline
\rowcolor[gray]{0.9} \textbf{Prompt} \\ \hline
\texttt{You are a budget allocation controller for a retrieval-augmented QA system.} \\
\texttt{Before retrieval starts, allocate integer retrieval budgets across all questions in the batch.} \\
\texttt{Each tier of retrieval has a known cost, and the downstream agent will spend the allocated budget on retrieval to gather evidence for answering.} \\
\texttt{Questions that likely need multi-hop, official, professional, or domain-specific evidence should receive more budget.} \\
\texttt{Questions that can likely be answered from the prompt or cheap evidence can receive less budget.} \\
\texttt{Return only the final tagged TSV allocation block.} \\
\texttt{Batch size: \textless batch\_size\textgreater{}} \\
\texttt{Total shared budget: \textless total\_budget\textgreater{}} \\
\texttt{Per-question bounds: min=\textless min\_budget\textgreater{}, max=\textless max\_text\textgreater{}} \\
\texttt{Retrieval tier options:} \\
\texttt{\textless tier\_desc\textgreater{}} \\
\texttt{Questions: \textless Questions\textgreater{}} \\
\texttt{Allocate budgets with these strict rules:} \\
\texttt{- Include every id exactly once.} \\
\texttt{- Each budget must be a non-negative integer within the per-question bounds.} \\
\texttt{- The sum of all budgets must equal the total shared budget.} \\
\texttt{- Budget is spent only by retrieval; unused budget may remain after the downstream agent answers.} \\
\texttt{Return the final allocation as plain text, not JSON. Use exactly this TSV block format:} \\
\texttt{BEGIN\_BUDGET\_ALLOCATION} \\
\texttt{\textless question id\textgreater{}  \textless integer budget\textgreater{}  \textless short reason\textgreater{}} \\
\texttt{...} \\
\texttt{END\_BUDGET\_ALLOCATION} \\
\texttt{Do not put Markdown tables, bullets, or JSON inside the final block.} \\
 \hline
\end{tabularx}
\caption{Prompt Template for Budget Allocation in Shared Budget setting.} 
\label{tab:agent_prompt_budget}
\end{table}

%% file: table/agent_casestudy_1_table.tex

\begin{table*}[t]
\centering
\small
\renewcommand\arraystretch{1.18}
\setlength{\fboxsep}{1.0pt}
\resizebox{\linewidth}{!}{
\begin{tabular}{p{0.88\textwidth}}
\hline
\rowcolor{gray!8}{\textbf{Case:} HotpotQA (ID: 5a70f39c5542994082a3e429). }\\
\hline
\textbf{Question:} What is the name of the independent candidate in Maine's 2010 gubernatorial race who finished ahead of Libby Mitchell?\\
\textbf{Ground-Truth:} \(\boxed{Eliot \;Cutler}\)\\
\textbf{Round 1:}
Okay, let's tackle this question.
The question is about a specific historical election, which likely requires external knowledge. \(\ldots\)
\textbf{\textcolor[rgb]{0.13,0.55,0.8}{The budget is 20, and I haven't spent anything yet.}}  \(\ldots\)
\textbf{\textcolor[rgb]{0.7,0.3,0.3}{Considering the tiers, tier 0 is free and includes sources like Wikipedia. That's the cheapest option. I should start by retrieving from tier 0. If the information isn't there, I can move to higher tiers.}} \(\ldots\)\\
\textbf{Action:} \textsc{Retrieve} \; \textsc{Tier}: 0 \; \textsc{Cost} : 0\\
\textbf{Round 2:}
The question is about 2010, and the context here is 2018. 
So I need to find out who the independent candidate was in 2010. \(\ldots\)
\textbf{\textcolor[rgb]{0.7,0.3,0.3}{Since the question is about a specific historical event, tier 0 (Wikipedia) is the best starting point.}} \textbf{\textcolor[rgb]{0.13,0.55,0.8}{The budget is 20.}} \(\ldots\)
Therefore, the agent should retrieve another passage from tier 0 to find the 2010 race details.  \(\ldots\)
\textbf{\textcolor[rgb]{0.13,0.55,0.8}{The budget allows for more retrievals.}}  \(\ldots\) \\
\textbf{Action: }  \textsc{Retrieve} \; \textsc{Tier}: 0 \; \textsc{Cost} : 0\\

\textbf{Round 3:}
The Passage 2 says: \(\ldots\) 
Therefore, Cutler would have more votes than Mitchell. 
Thus, the answer is Eliot Cutler.

\textbf{Action: } 
\textsc{Answer} \(\boxed{Eliot \; Cutler}\)\\
\textsc{Total Cost}: 0\; \textsc{Total Number of Passages} : 2\\
\hline
\end{tabular}}
\caption{\label{tab:agent-case-study}Case Study 1: the Agent Retrieves and Answers with \textbf{\textcolor[rgb]{0.7,0.3,0.3}{Tier Selection}}, and \textbf{\textcolor[rgb]{0.13,0.55,0.8}{Cost Consideration}}.}
\end{table*}

%% file: table/agent_casestudy_2_table.tex

\begin{table*}[t]
\centering
\small
\renewcommand\arraystretch{1.2}
\setlength{\fboxsep}{1.0pt}
\resizebox{\linewidth}{!}{
\begin{tabular}{p{0.88\textwidth}}
\hline
\rowcolor{gray!8}{\textbf{Case: } NQ (ID: -458159261891926107). }\\
\hline
\textbf{Question:} when was the last time mount ruapehu erupted?\\
\textbf{Ground-Truth:} \(\boxed{25\;September\;2007}\)\\
\textbf{Round 1:}
The question requires specific geological data about the volcano's eruptive history. 
I need to retrieve data from an external source. \(\ldots\)
\textbf{\textcolor[rgb]{0.13,0.55,0.8}{The budget is 20, and the cost for tier 0 is 0}}. \textbf{\textcolor[rgb]{0.7,0.3,0.3}{Tier 0 includes community resources like Wikipedia.}} \textbf{\textcolor[rgb]{0.13,0.55,0.8}{That's the cheapest option, so I should start there}}. If Wikipedia doesn't have the information, maybe I can move to a higher tier. \(\ldots\)\\
\textbf{Action:} \textsc{Retrieve} \; \textsc{Tier}: 0 \; \textsc{Cost} : 0\\


\textbf{Round 2:}
The current context 
mentions that Ruapehu \(\ldots\)
erupted between 15,000–2,000 years ago. 
But the question is about the last eruption, \(\ldots\)
which might be within the last few hundred years or even more recently. \(\ldots\)
The next step should be to retrieve another passage. 
\textbf{\textcolor[rgb]{0.13,0.55,0.8}{Since the budget is 20 and the first retrieval was from tier 0 (cost 0), there's still a lot left.}} Tier 0 is the cheapest, so using tier 0 again makes sense. 
\textbf{\textcolor[rgb]{0.7,0.3,0.3}{The reason is to check community resources like Wikipedia for the most recent eruption data. }}
If that doesn't have it, maybe move to tier 1 or 2, but tier 0 is the best first step. \\
\textbf{Action:} \textsc{Retrieve} \; \textsc{Tier}: 0 \; \textsc{Cost} : 0\\

\textbf{Round 3:}
\(\ldots\)
The information in the passages is historical but doesn't cover the most recent activity. 
\textbf{\textcolor[rgb]{0.7,0.3,0.3}{Looking at the tiers, Tier 0 is community resources like Wikipedia, which we've already used. 
Tier 1 is general open-web content like news sites, which might have more recent information. 
Tier 2 is official sources, which could be reliable but might be more expensive. }}
\textbf{\textcolor[rgb]{0.13,0.55,0.8}{The budget is 20, and the remaining is 20. Since we've already retrieved two passages from Tier 0, maybe we should try Tier 1 next. 
Let me check the cost: Tier 1 is 1 per passage. So retrieving from Tier 1 would cost 1, leaving 19. }}
Tier 1 is the next best option. \\
\textbf{Action:} \textsc{Retrieve} \; \textsc{Tier}: 1 \; \textsc{Cost} : 1\\

\textbf{Round 4:}
\(\ldots\)
Passage 3 is tier 1, but it's a general geology article and doesn't provide specific dates.\(\ldots\)
The user needs external knowledge here. 
\textbf{\textcolor[rgb]{0.13,0.55,0.8}{The budget is 20, spent 1, remaining 19.}} 
Since tier 0 was used already \(\ldots\)
\textbf{\textcolor[rgb]{0.7,0.3,0.3}{tier 1 might have more recent info.}}
If I choose tier 1 again, cost 1, which is possible.\(\ldots\)\\
\textbf{Action:} \textsc{Retrieve} \; \textsc{Tier}: 1 \; \textsc{Cost} : 1\\

\textbf{Round 5:}
\(\ldots\)
Passage 4 talks about \(\ldots\) not Ruapehu.
I need to retrieve more information. 
Therefore, the next action should be to retrieve another passage, possibly from tier 1 or maybe tier 2. \(\ldots\)
But given the budget, if I retrieve from tier 1 again, \textbf{\textcolor[rgb]{0.13,0.55,0.8}{it's 1 cost, which is affordable.}} \(\ldots\)
However, if the information isn't found there, then tier 2 might be needed. 
The next tier to consider is tier 1 again, as it's the next most appropriate after tier 0. \(\ldots\)\\
\textbf{Action:} \textsc{Retrieve} \; \textsc{Tier}: 1 \; \textsc{Cost} : 1\\

\textbf{Round 6:}
\(\ldots\)
Passages 3, 4, and 5 are from tier 1, but they talk about other volcanoes \(\ldots\)
\textbf{\textcolor[rgb]{0.13,0.55,0.8}{Since the budget allows for more retrievals (remaining is 17)}}, I need to decide whether to retrieve another passage. Tier 0 is cheaper, but maybe tier 1 has more recent news articles. However, the previous retrievals from tier 1 didn't provide the answer. \textbf{\textcolor[rgb]{0.7,0.3,0.3}{Maybe I should check a higher tier, like tier 2, but that's more expensive. }}
\textbf{\textcolor[rgb]{0.13,0.55,0.8}{Wait, the user's remaining budget is 17, and tier 2 costs 4 per passage. If I retrieve from tier 2, I could spend 4, leaving 13.}} \(\ldots\)
The user's previous steps have already tried tier 0 and 1 without success.\\
Since the budget allows, perhaps it's worth trying tier 2. \\
\textbf{Action:} \textsc{Retrieve} \; \textsc{Tier}: 2 \; \textsc{Cost} : 4\\

\textbf{Round 7:}
\(\ldots\)
None of these passages provide the needed information.
\textbf{\textcolor[rgb]{0.13,0.55,0.8}{The user's trajectory shows that they've already spent 7 out of 20 budget points, with 13 remaining. }}
\textbf{\textcolor[rgb]{0.7,0.3,0.3}{They've tried tiers 0, 1, and 2. }}
Since the current context doesn't have the required information, the next step is to retrieve another passage.
\(\ldots\)
Wait, the user has already used tier 2 once. But maybe they need another source from tier 0 or 1. 
\textbf{\textcolor[rgb]{0.13,0.55,0.8}{The remaining budget is 13, so even if they retrieve another tier 1 (cost 1), they can do that. }}
\(\ldots\)
Since the user has already used tier 0 and 1, maybe tier 2 again? \textbf{\textcolor[rgb]{0.13,0.55,0.8}{But the cost is 4, and remaining budget is 13. If they retrieve tier 2 again, they can spend 4, leaving 9. }}\(\ldots\)
\textbf{\textcolor[rgb]{0.13,0.55,0.8}{Since tier 0 is the cheapest, perhaps retrieve another tier 0 passage. }}
But the user has already retrieved two tier 0 passages. 
However, the system allows retrieving from the same tier multiple times. So the next action is to retrieve from tier 0 again, hoping to find the answer.\\
\textbf{Action:} \textsc{Retrieve} \; \textsc{Tier}: 0 \; \textsc{Cost} : 0\\

\textbf{Round 8:}
\(\ldots\)
Wait, actually, in Passage 7 (tier 0), there's a direct answer: it last erupted on the 25th of September 2007. That seems specific. \(\ldots\)
Yes, in the current context, Passage 7 is listed. So the answer is September 25, 2007. So the correct action is to answer now.\\
\textbf{Action: } \textsc{Answer} \(\boxed{2007}\)\\
\textsc{Total Cost}: 7\; \textsc{Total Number of Passages} : 7\\
\hline
\end{tabular}}
\caption{\label{tab:agent-case-study-2}Case Study 2: the Agent Retrieves and Answers with \textbf{\textcolor[rgb]{0.7,0.3,0.3}{Tier Selection}}, and \textbf{\textcolor[rgb]{0.13,0.55,0.8}{Cost Consideration}}.}
\end{table*}

%% file: table/agent_casestudy_3_table.tex

\begin{table*}[t]
\centering
\small
\renewcommand\arraystretch{1.18}
\setlength{\fboxsep}{1.0pt}
\resizebox{\linewidth}{!}{
\begin{tabular}{p{0.88\textwidth}}
\hline
\rowcolor{gray!8}{\textbf{Case:} MMLU (ID: 1022). }\\
\hline
\textbf{Question:} A 1-year-old male is brought to the emergency department with blistered fresh burns over his hands and feet in a stocking and glove distribution. His mother reports that he was accidentally burned when he turned on the hot water in the bathtub while she was answering the telephone in the bedroom. In addition to debridement of the burns, pain management, a thorough history and physical examination, and baseline laboratory studies, the most appropriate next step is to Options: A. administer a 5\% albumin infusion B. administer intravenous steroids C. contact child protective services D. discharge the patient home with his parents\\
\textbf{Ground-Truth:} \(\boxed{C}\)\\
\textbf{Round 1:}
Maybe I need to check if there's any standard protocol here. \(\ldots\)
\textbf{\textcolor[rgb]{0.7,0.3,0.3}{So, the agent should retrieve from tier 2, which includes medical sources.}} 
So the next action is to retrieve from tier 2 to get the appropriate guidelines.\\ 
\textbf{Action:} \textsc{Retrieve} \; \textsc{Tier}: 2 \; \textsc{Cost} : 4\\
\textbf{Round 2:}
The options include contacting child protective services (option C), which might be relevant because the burn pattern is suspicious for abuse \(\ldots\).
Maybe the answer is C, but the current context doesn't support that. 
Therefore, I need to retrieve another passage. \\
\textbf{\textcolor[rgb]{0.13,0.55,0.8}{But the remaining budget is 16. 
The next step would be to retrieve another passage from tier 2, which costs 4, leaving 12. }} \\
\textbf{Action:} \textsc{Retrieve} \; \textsc{Tier}: 2 \; \textsc{Cost} : 4\\

\textbf{Round 3:}
\(\ldots\) 
Passage 2 says that \(\ldots\) 
The passages provided don't mention CPS or child abuse protocols. 
But the answer might be C. 
But without retrieval, can't be sure. 
\textbf{\textcolor[rgb]{0.13,0.55,0.8}{The budget allows for more retrievals (remaining is 12),}} 
\textbf{\textcolor[rgb]{0.7,0.3,0.3}{I should check if Tier 2 has information on this. }}
\textbf{\textcolor[rgb]{0.13,0.55,0.8}{If I need to retrieve another passage from Tier 2, that would cost 4, leaving 8.}}
\\
\textbf{Action:} \textsc{Retrieve} \; \textsc{Tier}: 2 \; \textsc{Cost} : 4\\

\textbf{Round 4:}
The options include contacting CPS, which would be relevant if there's suspicion of abuse.
Passage 2 says that burns \(\ldots\) 
So maybe the answer requires external knowledge.
\textbf{\textcolor[rgb]{0.13,0.55,0.8}{The user has already spent 12, so remaining is 8. }}
\textbf{\textcolor[rgb]{0.13,0.55,0.8}{Tier 2 costs 4 per passage. }}
\textbf{\textcolor[rgb]{0.13,0.55,0.8}{So if I retrieve another tier 2 passage, that would cost 4, leaving 4}}
\\
\textbf{Action:} \textsc{Retrieve} \; \textsc{Tier}: 2 \; \textsc{Cost} : 4\\

\textbf{Round 5:}
In young children, 
such burns can also be signs of neglect or abuse. 
Passage 1 and 2 talk about  \(\ldots\).
Passage 3 and 4 mention \(\ldots\).
However, none of the passages directly address child protective services (CPS) or suspicion of abuse.
\textbf{\textcolor[rgb]{0.13,0.55,0.8}{But the remaining budget is 4, and tier 2 costs 4. 
If I retrieve another passage from tier 2, that would use up the remaining budget. }}
\textbf{\textcolor[rgb]{0.7,0.3,0.3}{So, even though the budget is low, maybe I should retrieve one more passage from tier 2 to get that info.}}
Then, after that, the agent can answer based on the new passage. \\
\textbf{Action:} \textsc{Retrieve} \; \textsc{Tier}: 2 \; \textsc{Cost} : 4\\

\textbf{Round 6:}
However, none of the passages specifically mentions child abuse suspicion or CPS protocols.
\textbf{\textcolor[rgb]{0.7,0.3,0.3}{The passages retrieved are from tier 2, which are medical sources. }}
Therefore, the answer might require external knowledge beyond the given context. 
\textbf{\textcolor[rgb]{0.13,0.55,0.8}{But the budget is already exhausted (remaining 0), so I can't retrieve more. }}
However, if the context doesn't provide that, but the model is supposed to answer based on existing knowledge, then answer C\\
\textbf{Action: } \textsc{Answer} \(\boxed{C}\)\\
\textsc{Total Cost}: 20\; \textsc{Total Number of Passages} : 5\\
\hline
\end{tabular}}
\caption{\label{tab:agent-case-study-3}Case Study 3: the Agent Retrieves and Answers with \textbf{\textcolor[rgb]{0.7,0.3,0.3}{Tier Selection}}, and \textbf{\textcolor[rgb]{0.13,0.55,0.8}{Cost Consideration}}.}
\end{table*}

%% file: table/agent_casestudy_4_table.tex

\begin{table*}[t]
\centering
\small
\renewcommand\arraystretch{1.18}
\setlength{\fboxsep}{1.0pt}
\resizebox{\linewidth}{!}{
\begin{tabular}{p{0.88\textwidth}}
\hline
\rowcolor{gray!8}{\textbf{Case:} HotpotQA. }\\
\hline
\textbf{Q1:} Did Lewis Allen or Danny DeVito play Louie De Palma?\\
\textbf{Q2:} Who wrote Tom Vaughan's popular 2008 film?\\
\textbf{Q3:} Are Ruggero Deodato from Italy, and Mexican Alejandro Springall, both film directors?\\
\textbf{Q4:} Which actor and film director is from German Jewish descent, Kurt Gerron or Dell Henderson?\\
\textbf{Q5:} Which genus has more species, Monstera or Cercis?\\
\textbf{Responce:}\\
\textbf{Question 1:} \(\ldots\)
\textbf{\textcolor[rgb]{0.7,0.3,0.3}{the information is likely to be found in public knowledge or general open-web content (tier 1). }}
So maybe tier 1, budget allocation of 1 or 2. 
perhaps tier 1 with budget 1.
\textbf{Question 2: }\(\ldots\)
\textbf{\textcolor[rgb]{0.7,0.3,0.3}{this information might be found in general open-web sources (tier 1). }}
So again, tier 1, budget 1 or 2. Maybe 2.
\textbf{Question 3:} 
\(\ldots\)
\textbf{\textcolor[rgb]{0.7,0.3,0.3}{This is likely to be found in tier 1 (general open-web content) or tier 2 if it's more specific. }}
maybe tier 1. Budget 1.
\textbf{Question 4: } \(\ldots\)
\textbf{\textcolor[rgb]{0.7,0.3,0.3}{This is a specific historical fact, which might require more detailed sources. 
If the answer isn't readily available in general sources, maybe tier 2. 
But if it's public knowledge, tier 1. }}
Let's say tier 2, budget 4.
Question 5: \(\ldots\)
\textbf{\textcolor[rgb]{0.7,0.3,0.3}{It might require more specific scientific sources}}, which could be tier 2. So tier 2, budget 4.
Now, let's try to assign budgets. 
Let's assume that questions 1, 2, 3 are tier 1 (budget 1 each), question 4 tier 2 (budget 4), question 5 tier 2 (budget 4).
\textbf{\textcolor[rgb]{0.13,0.55,0.8}{Total would be 1+1+1+4+4=11}}, which is way under 100. So clearly, that's not sufficient. 

So I need to assign each question a budget such that the sum is 100. 

Let me reassess each question again with the idea of what's needed:
\textbf{\textcolor[rgb]{0.13,0.55,0.8}{But that gives 1+2+1+4+4=12}}. Still way too low. 
Let me think again about the questions: \(\ldots\)
4), question 5 tier 2 (budget 4).
\textbf{\textcolor[rgb]{0.13,0.55,0.8}{Total is 20+20+10+40+20=110}}. Over the limit.
Alternatively, adjust: \(\ldots\) 
4), question 5 tier 2 (budget 4).
\textbf{\textcolor[rgb]{0.13,0.55,0.8}{Total is 15+15+10+30+20=90}}. Need 10 more. 
So distribute the remaining 10 across the questions. Maybe increase tier 2 budgets: Question 4: 40, question 5: 20. \textbf{\textcolor[rgb]{0.13,0.55,0.8}{Total 15+15+10+40+20=100}}. That works.

But I need to check if the reasoning for tiers is correct. \(\ldots\)
Alternatively, \textbf{\textcolor[rgb]{0.7,0.3,0.3}{maybe question 5 is in tier1 because it's a general botanical fact. }}
But if I change question 5 to tier 1, then the total becomes: \textbf{\textcolor[rgb]{0.13,0.55,0.8}{Question1:15, q2:15, q3:10, q4:40, q5:20 (total 100)}}. But if question 5 is tier1, then budget 20 (assuming tier1). 
But I need to check if the other questions need higher tiers.
Alternatively, maybe question 4 is tier2, and question 5 is tier2. 
If question5 is tier2, then budget for q5 is 20 (which is 5 passages at 4 per passage). \textbf{\textcolor[rgb]{0.13,0.55,0.8}{Then total would be:q1:15, q2:15, q3:10, q4:40, q5:20.}} Total 100. That works. But maybe other allocations are possible. Let me think if any of the questions require tier0. 

Let's assume that some questions can be answered with tier0 (cost 0). 
For example: Question 3: \(\ldots\)
\textbf{\textcolor[rgb]{0.7,0.3,0.3}{This is factual information. If the answer is in public knowledge (Wikipedia), then tier 0}}. So allocate 0. But that would save budget. 
So perhaps: Question3: tier0, budget 0. 
So total would be 4), question 5 tier 2 (budget 4).
\textbf{\textcolor[rgb]{0.13,0.55,0.8}{15+15+0+40+20=90}}. Need 10 more. 
\textbf{\textcolor[rgb]{0.13,0.55,0.8}{Let's add 10 to question 1: 25, q2:15, q3:0, q4:40, q5:20. Total 25+15+0+40+20=100}}.

\textbf{Final Allocation: } \(\boxed{Q1: 25, Q2:15, Q3:0, Q4:40, Q5:20.}\)\\
\hline
\end{tabular}}
\caption{\label{tab:agent-case-study-4}Case Study 4: Flexible Budget Allocation in Shared Budget Setting \textbf{\textcolor[rgb]{0.7,0.3,0.3}{Tier Selection}}, and \textbf{\textcolor[rgb]{0.13,0.55,0.8}{Cost Consideration}.}
}
\end{table*}

%% file: table/agent_casestudy_5_medqa_qwen_table.tex


\begin{table*}[t]
\centering
\small
\renewcommand\arraystretch{1.18}
\setlength{\fboxsep}{1.0pt}
\resizebox{\textwidth}{!}{
\begin{tabular}{p{0.88\textwidth}}
\hline
\rowcolor{gray!8}{\textbf{Case} MedQA ( ID: 563) }\\
\hline
\textbf{Question:} A 47-year-old woman seeks evaluation at your office because she has had postcoital vaginal bleeding for the past 8 months with occasional intermenstrual watery, blood-tinged vaginal discharge. Her family history is negative for malignancies and inherited disorders. She is the result of a pregnancy complicated by numerous miscarriages in the 1960s, for which her mother received diethylstilbestrol. During a pelvic examination, you notice a polypoid mass on the anterior wall of the vagina. The bimanual examination is negative for adnexal masses. You suspect the presence of carcinoma and, therefore, send tissue samples to pathology, which confirmed the presence of malignant cells. Which of the following is the most likely malignant tumor in this patient? \\
Options: "A": "Clear cell adenocarcinoma", "B": "Melanoma", "C": "Botryoid sarcoma", "D": "Verrucous carcinoma" \\
\textbf{Ground-Truth:} \(\boxed{A}\)\\
\textbf{Round 1:}\\
Since DES is a known risk factor for clear cell adenocarcinoma \(\ldots\)
Since I'm not 100\% certain without checking, maybe I should retrieve. 
But the budget allows for it. 
\textbf{\textcolor[rgb]{0.7,0.3,0.3}{Tier 2 is for medical sources, which would be the best for this. 
So, I should retrieve from tier 2 to confirm the association between DES and clear cell adenocarcinoma.}}
\textbf{\textcolor[rgb]{0.13,0.55,0.8}{But the remaining budget is 40, and tier 2 costs 4 per passage. }} \\
\textbf{Action:} \textsc{Retrieve} \; \textsc{Tier}: 2 \; \textsc{Cost} : 4\\

\textbf{Round 2:}\\
\(\ldots\) Since the current context already provides the necessary information \(\ldots\) 
there's no need for further retrieval. 
The answer can be confidently given as A. 
The budget remaining is 36, which is sufficient, but since the info is already there, answering now is appropriate.\\
\textbf{Action: } \textsc{Answer} \(\boxed{A}\)\\
\textsc{Total Cost}: 4\; \textsc{Total Number of Passages} : 1\\
\hline
\end{tabular}}
\caption{\label{tab:agent-case-study-medqa-qwen}Case Study 5: the Agent (Qwen3-8B, Budget=40) Retrieves and Answers with \textbf{\textcolor[rgb]{0.7,0.3,0.3}{Tier Selection}}, and \textbf{\textcolor[rgb]{0.13,0.55,0.8}{Cost Consideration}}.}
\end{table*}

%% file: table/agent_casestudy_6_medqa_llama_table.tex


\begin{table*}[t]
\centering
\small
\renewcommand\arraystretch{1.18}
\setlength{\fboxsep}{1.0pt}
\resizebox{\textwidth}{!}{
\begin{tabular}{p{0.88\textwidth}}
\hline
\rowcolor{gray!8}{\textbf{Case} MedQA ( ID: 563) }\\
\hline
\textbf{Question:} A 47-year-old woman seeks evaluation at your office because she has had postcoital vaginal bleeding for the past 8 months with occasional intermenstrual watery, blood-tinged vaginal discharge. Her family history is negative for malignancies and inherited disorders. She is the result of a pregnancy complicated by numerous miscarriages in the 1960s, for which her mother received diethylstilbestrol. During a pelvic examination, you notice a polypoid mass on the anterior wall of the vagina. The bimanual examination is negative for adnexal masses. You suspect the presence of carcinoma and, therefore, send tissue samples to pathology, which confirmed the presence of malignant cells. Which of the following is the most likely malignant tumor in this patient? \\
Options: "A": "Clear cell adenocarcinoma", "B": "Melanoma", "C": "Botryoid sarcoma", "D": "Verrucous carcinoma" \\
\textbf{Ground-Truth:} \(\boxed{A}\)\\
\textbf{Round 1 -- 2:}\\
Question requires medical knowledge, so a higher tier is preferred for accuracy.\\
\textbf{Action:} \textsc{Retrieve} \; \textsc{Tier}: 2 \; \textsc{Cost} : 4\\


\textbf{Round 3:}\\
Since the current context is not sufficient to confidently answer the question, and \textbf{\textcolor[rgb]{0.13,0.55,0.8}{the remaining budget is more than the cost of a passage from tier 2}}, the next action should be to \textbf{\textcolor[rgb]{0.7,0.3,0.3}{retrieve a passage from tier 2 to gather more specific information about the possible malignant tumors in the patient}}.\\
The question asks for the most likely malignant tumor in the patient, suggesting that more specific information is needed, and a higher tier is preferred for accuracy.\\
\textbf{Action:} \textsc{Retrieve} \; \textsc{Tier}: 2 \; \textsc{Cost} : 4\\

\textbf{Round 4 -- 8:}\\
The question asks for the most likely malignant tumor in the patient, suggesting that more specific information is needed, and a higher tier is preferred for accuracy.\\
\textbf{Action:} \textsc{Retrieve} \; \textsc{Tier}: 2 \; \textsc{Cost} : 4\\





\textbf{Round 9:}\\
Based on the current context and the question asked, it is clear that the question requires outside knowledge. 
\textbf{\textcolor[rgb]{0.13,0.55,0.8}{Since the remaining budget is 8, and the cost of tier 2 is 4, we can afford to retrieve one more passage.}} Therefore, the next action is to retrieve one passage from tier 2.\\
\textbf{Action:} \textsc{Retrieve} \; \textsc{Tier}: 2 \; \textsc{Cost} : 4\\

\textbf{Round 10:}
The question asks \(\ldots\)
suggesting that more specific information is needed, and a higher tier is preferred for accuracy.\\
\textbf{Action:} \textsc{Retrieve} \; \textsc{Tier}: 2 \; \textsc{Cost} : 4\\

\textbf{Round 11:}
\textbf{Action: } \textsc{Answer} \(\boxed{A}\)\\
\textsc{Total Cost}: 40\; \textsc{Total Number of Passages} : 10\\
\hline
\end{tabular}}
\caption{\label{tab:agent-case-study-medqa-llama}Case Study 6: the Agent (Llama3.1-8B-Instruct, Budget=40) Retrieves and Answers with \textbf{\textcolor[rgb]{0.7,0.3,0.3}{Tier Selection}}, and \textbf{\textcolor[rgb]{0.13,0.55,0.8}{Cost Consideration}}.}
\end{table*}

%% file: figure_tex/tier_distribution_full.tex
\begin{figure*}[t] 
\centering
    \includegraphics[width=1.0\textwidth]{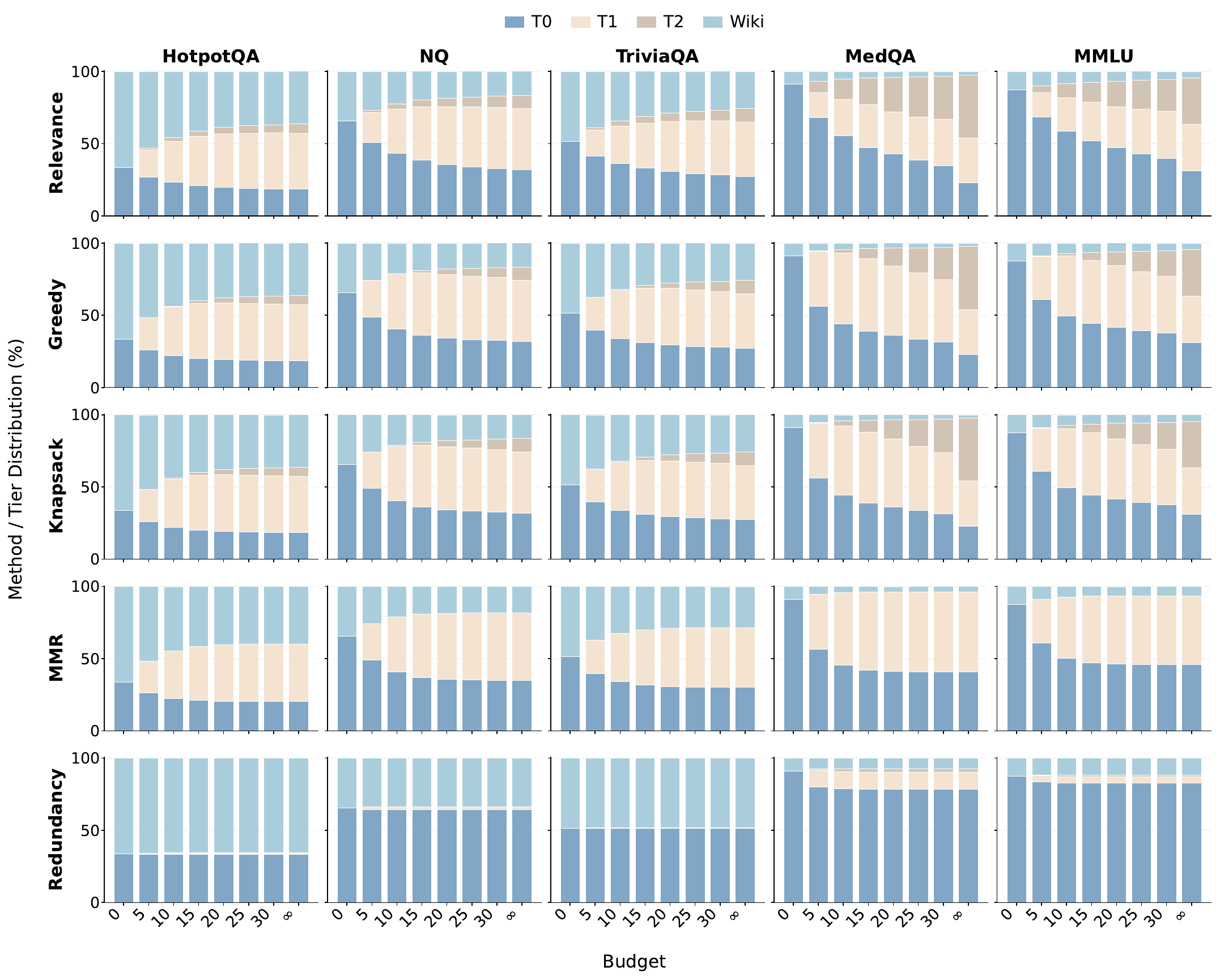}
    \caption{Full Tier Distribution. Wiki belongs to T0. We conduct analysis experiments using the Qwen3-8B model. The observation in Llama3.1-8B-instruct is the same, since the passages are selected by rule-based selectors.} \label{fig:full_tier_distribution}
\end{figure*}

%% file: table/overall_performance_qwen_ci.tex
\begin{table*}[htbp]
\centering
\scriptsize
\setlength{\tabcolsep}{1.2pt} 
\resizebox{\linewidth}{!}{
\begin{tabular}{ll cc cc cc cc cc}
\toprule
\multirow{3}{*}{\textbf{Method}} & \multirow{3}{*}{\textbf{Bgt/k}} 
& \multicolumn{6}{c}{\textbf{Open-Domain Datasets}} 
& \multicolumn{4}{c}{\textbf{Domain-Specific Datasets}} \\
\cmidrule(lr){3-8} \cmidrule(lr){9-12}
& & \multicolumn{2}{c}{\textbf{HotpotQA}} 
& \multicolumn{2}{c}{\textbf{NQ}} 
& \multicolumn{2}{c}{\textbf{TriviaQA}} 
& \multicolumn{2}{c}{\textbf{MedQA}} 
& \multicolumn{2}{c}{\textbf{MMLU}} \\
\cmidrule(lr){3-4} \cmidrule(lr){5-6} \cmidrule(lr){7-8} \cmidrule(lr){9-10} \cmidrule(lr){11-12}
& & \textbf{C} & \textbf{F1 (\%)} 
& \textbf{C} & \textbf{F1 (\%)} 
& \textbf{C} & \textbf{F1 (\%)} 
& \textbf{C} & \textbf{F1 (\%)} 
& \textbf{C} & \textbf{F1 (\%)} \\
Vanilla LLM & - & \textsc{N/A} & $37.80_{\pm 8.66}$ & \textsc{N/A} & $19.81_{\pm 6.54}$ & \textsc{N/A} & $43.57_{\pm 8.75}$ & \textsc{N/A} & $45.00_{\pm 9.80}$ & \textsc{N/A} & $75.00_{\pm 8.53}$ \\
\midrule
\multirow{6}{*}{Top-k} 
  & 5 & \textsc{N/A} & $31.90_{\pm 8.40}$ & \textsc{N/A} & $37.14_{\pm 7.75}$ & \textsc{N/A} & $56.74_{\pm 8.85}$ & \textsc{N/A} & $66.00_{\pm 9.33}$ & \textsc{N/A} & $83.00_{\pm 7.40}$ \\
 & 10 & \textsc{N/A} & $37.19_{\pm 8.61}$ & \textsc{N/A} & $36.87_{\pm 7.88}$ & \textsc{N/A} & $54.16_{\pm 8.67}$ & \textsc{N/A} & $72.00_{\pm 8.84}$ & \textsc{N/A} & $84.00_{\pm 7.22}$ \\
 & 15 & \textsc{N/A} & $34.45_{\pm 8.71}$ & \textsc{N/A} & $40.10_{\pm 8.28}$ & \textsc{N/A} & $56.48_{\pm 8.43}$ & \textsc{N/A} & $68.00_{\pm 9.19}$ & \textsc{N/A} & $82.00_{\pm 7.57}$ \\
 & 20 & \textsc{N/A} & $33.02_{\pm 8.56}$ & \textsc{N/A} & $39.06_{\pm 8.07}$ & \textsc{N/A} & $59.41_{\pm 8.45}$ & \textsc{N/A} & $62.00_{\pm 9.56}$ & \textsc{N/A} & $83.00_{\pm 7.36}$ \\
 & 25 & \textsc{N/A} & $37.70_{\pm 8.72}$ & \textsc{N/A} & $39.60_{\pm 8.21}$ & \textsc{N/A} & $60.30_{\pm 8.52}$ & \textsc{N/A} & $61.00_{\pm 9.61}$ & \textsc{N/A} & $80.00_{\pm 7.88}$ \\
 & 30 & \textsc{N/A} & $35.43_{\pm 8.80}$ & \textsc{N/A} & $38.25_{\pm 8.07}$ & \textsc{N/A} & $63.11_{\pm 8.37}$ & \textsc{N/A} & $64.00_{\pm 9.46}$ & \textsc{N/A} & $78.00_{\pm 8.16}$ \\
\midrule
\multirow{8}{*}{Relevance} 
 &        0  & 0.00 & $34.44_{\pm 8.55}$ & 0.00 & $37.49_{\pm 7.91}$ & 0.00 & $55.37_{\pm 8.49}$ & 0.00 & $67.00_{\pm 9.26}$ & 0.00 & $80.00_{\pm 7.88}$ \\
 &        5  & 4.80 & $33.73_{\pm 8.47}$ & 5.00 & $42.54_{\pm 8.03}$ & 4.90 & $59.39_{\pm 8.58}$ & 5.00 & $67.00_{\pm 9.26}$ & 5.00 & $84.00_{\pm 7.22}$ \\
 &        10 & 9.20 & $35.73_{\pm 8.75}$ & 9.60 & $42.95_{\pm 8.44}$ & 9.10 & $58.28_{\pm 8.65}$ & 10.0 & $65.00_{\pm 9.40}$ & 9.90 & $80.00_{\pm 7.88}$ \\
 &        15 & 12.9 & $35.35_{\pm 8.75}$ & 13.6 & $42.41_{\pm 8.22}$ & 12.6 & $59.30_{\pm 8.60}$ & 14.9 & $66.00_{\pm 9.33}$ & 14.7 & $82.00_{\pm 7.57}$ \\
 &        20 & 15.5 & $36.44_{\pm 8.67}$ & 16.8 & $40.42_{\pm 8.15}$ & 15.6 & $57.48_{\pm 8.59}$ & 19.9 & $68.00_{\pm 9.19}$ & 19.6 & $81.00_{\pm 7.73}$ \\
 &        25 & 17.2 & $38.70_{\pm 8.79}$ & 19.1 & $42.51_{\pm 8.05}$ & 18.0 & $62.79_{\pm 8.58}$ & 25.0 & $67.00_{\pm 9.26}$ & 24.1 & $80.00_{\pm 7.88}$ \\
 &        30 & 18.1 & $38.11_{\pm 8.80}$ & 20.9 & $36.50_{\pm 7.78}$ & 19.7 & $59.99_{\pm 8.41}$ & 29.8 & $64.00_{\pm 9.46}$ & 28.1 & $79.00_{\pm 8.02}$ \\
 & $\infty$  & 19.2 & $35.49_{\pm 8.70}$ & 23.4 & $39.50_{\pm 8.09}$ & 22.3 & $64.43_{\pm 8.35}$  & 61.7  & $67.00_{\pm 9.26}$& 48.2 & $80.00_{\pm 7.88}$ \\
\midrule
\multirow{8}{*}{Greedy} 
 & 0  & 0.00 & $34.98_{\pm 8.39}$ & 0.00 & $35.76_{\pm 7.79}$ & 0.00 & $55.44_{\pm 8.49}$ & 0.00 & $61.00_{\pm 9.61}$ & 0.00 & $80.00_{\pm 7.88}$ \\
 & 5  & 4.70 & $33.40_{\pm 8.31}$ & 4.90 & $41.79_{\pm 8.14}$ & 4.80 & $59.86_{\pm 8.55}$ & 4.80 & $65.00_{\pm 9.40}$ & 4.80 & $82.00_{\pm 7.57}$ \\
 & 10 & 8.90 & $39.64_{\pm 8.85}$ & 9.40 & $41.13_{\pm 8.29}$ & 8.80 & $57.96_{\pm 8.76}$ & 9.20 & $66.00_{\pm 9.33}$ & 9.10 & $84.00_{\pm 7.22}$ \\
 & 15 & 12.6 & $40.90_{\pm 9.00}$ & 12.9 & $45.19_{\pm 8.46}$ & 12.0 & $59.90_{\pm 8.49}$ & 13.7 & $64.00_{\pm 9.46}$ & 13.7 & $79.00_{\pm 8.02}$ \\
 & 20 & 15.0 & $36.54_{\pm 8.77}$ & 16.1 & $40.28_{\pm 8.20}$ & 14.9 & $59.97_{\pm 8.59}$ & 18.5 & $62.00_{\pm 9.56}$ & 18.2 & $82.00_{\pm 7.57}$ \\
 & 25 & 16.9 & $35.55_{\pm 8.62}$ & 18.7 & $40.08_{\pm 8.16}$ & 17.6 & $59.83_{\pm 8.45}$ & 23.5 & $66.00_{\pm 9.33}$ & 23.0 & $84.00_{\pm 7.22}$ \\
 & 30 & 18.0 & $35.26_{\pm 8.80}$ & 20.5 & $43.09_{\pm 7.94}$ & 19.4 & $61.79_{\pm 8.43}$ & 28.6 & $69.00_{\pm 9.11}$ & 27.1 & $78.00_{\pm 8.16}$ \\
 & $\infty$ &19.2 & $34.57_{\pm 8.69}$ & 23.4 & $40.66_{\pm 8.14}$ &  22.3 & $60.78_{\pm 8.48}$ & 61.7 & $64.00_{\pm 9.46}$  & 48.2 & $79.00_{\pm 8.02}$ \\
\midrule
\multirow{8}{*}{Knapsack} 
 &       0  & 0.00 & $36.31_{\pm 8.77}$ & 0.00 & $37.11_{\pm 7.91}$ & 0.00 & $55.57_{\pm 8.50}$ & 0.00 & $62.00_{\pm 9.56}$ & 0.00 & $81.00_{\pm 7.73}$ \\
 &       5  & 4.80 & $33.70_{\pm 8.49}$ & 5.00 & $43.25_{\pm 8.28}$ & 4.80 & $58.94_{\pm 8.49}$ & 4.90 & $66.00_{\pm 9.33}$ & 4.90 & $83.00_{\pm 7.40}$ \\
 &       10 & 9.20 & $40.33_{\pm 8.84}$ & 9.50 & $41.64_{\pm 8.02}$ & 8.90 & $60.68_{\pm 8.52}$ & 9.60 & $65.00_{\pm 9.40}$ & 9.60 & $82.00_{\pm 7.57}$ \\
 &       15 & 12.6 & $35.02_{\pm 8.76}$ & 13.2 & $40.67_{\pm 8.29}$ & 12.3 & $60.11_{\pm 8.56}$ & 14.3 & $67.00_{\pm 9.26}$ & 14.1 & $82.00_{\pm 7.57}$ \\
 &       20 & 15.3 & $37.09_{\pm 8.74}$ & 16.4 & $41.41_{\pm 8.10}$ & 15.5 & $57.52_{\pm 8.65}$ & 19.2 & $63.00_{\pm 9.51}$ & 19.0 & $77.00_{\pm 8.29}$ \\
 &       25 & 17.1 & $34.97_{\pm 8.72}$ & 18.9 & $40.84_{\pm 8.12}$ & 17.8 & $61.24_{\pm 8.45}$ & 24.3 & $65.00_{\pm 9.40}$ & 23.6 & $79.00_{\pm 8.02}$ \\
 &       30 & 18.0 & $36.93_{\pm 8.79}$ & 20.7 & $43.32_{\pm 8.03}$ & 19.6 & $62.80_{\pm 8.28}$ & 29.3 & $64.00_{\pm 9.46}$ & 27.7 & $78.00_{\pm 8.16}$ \\
 & $\infty$ & 19.2 & $35.18_{\pm 8.58}$ & 23.4 & $39.34_{\pm 7.92}$ & 22.3 & $62.87_{\pm 8.32}$ & 61.7 & $61.00_{\pm 9.61}$ & 48.2 & $76.00_{\pm 8.41}$ \\
\midrule
\multirow{8}{*}{MMR} 
 &       0  & 0.00 & $34.49_{\pm 8.49}$ & 0.00 & $38.92_{\pm 8.04}$  & 0.00 & $53.71_{\pm 8.50}$   & 0.00 &  $64.00_{\pm 9.46}$ & 0.00 & $81.00_{\pm 7.73}$ \\
 &       5  & 4.70 & $34.63_{\pm 8.56}$ & 4.90 & $42.55_{\pm 8.17}$  & 4.70 & $60.38_{\pm 8.56}$  & 4.70 &  $65.00_{\pm 9.40}$  & 4.60 & $83.00_{\pm 7.40}$ \\
 &       10 & 8.00 & $37.17_{\pm 8.76}$ & 8.80 & $44.11_{\pm 8.18}$  & 8.00 & $62.89_{\pm 8.13}$  & 7.60 &  $66.00_{\pm 9.33}$  & 7.90 & $80.00_{\pm 7.88}$ \\
 &       15 & 9.80 & $37.92_{\pm 8.90}$ & 11.1 & $41.80_{\pm 8.24}$  & 9.90 & $58.26_{\pm 8.61}$  & 8.80 &  $63.00_{\pm 9.51}$  & 9.10 & $84.00_{\pm 7.22}$ \\
 &       20 & 10.6 & $40.77_{\pm 9.00}$ & 12.0 & $43.67_{\pm 8.29}$  & 10.9 & $58.87_{\pm 8.57}$  & 9.20 &  $65.00_{\pm 9.40}$  & 9.50 & $79.00_{\pm 8.02}$ \\
 &       25 & 10.9 & $39.59_{\pm 8.91}$ & 12.5 & $43.81_{\pm 8.22}$  & 11.2 & $59.94_{\pm 8.49}$  & 9.30 &  $67.00_{\pm 9.26}$  & 9.60 & $84.00_{\pm 7.22}$ \\
 &       30 & 10.9 & $38.35_{\pm 8.77}$ & 12.5 & $41.34_{\pm 8.11}$  & 11.2 & $59.71_{\pm 8.50}$  & 9.40 &  $65.00_{\pm 9.40}$  & 9.70 & $78.00_{\pm 8.16}$ \\
 & $\infty$ & 10.9 & $37.71_{\pm 8.95}$ & 12.5 & $42.24_{\pm 8.33}$  & 11.2 & $59.30_{\pm 8.57}$  & 9.40 &  $70.00_{\pm 9.03}$  & 9.70 & $80.00_{\pm 7.88}$ \\
\midrule
\multirow{8}{*}{Redundancy} 
 & 0        & 0.00 & $32.51_{\pm 8.25}$ & 0.00 & $36.51_{\pm 7.95}$ & 0.00 & $55.28_{\pm 8.64}$ & 0.00 & $63.00_{\pm 9.51}$ & 0.00 & $83.00_{\pm 7.40}$ \\
 & 5        & 0.20 & $32.61_{\pm 8.26}$ & 0.30 & $37.40_{\pm 8.02}$ & 0.10 & $57.47_{\pm 8.50}$ & 1.20 & $69.00_{\pm 9.11}$ & 0.60 & $80.00_{\pm 7.88}$ \\
 & 10       & 0.20 & $33.34_{\pm 8.41}$ & 0.30 & $39.90_{\pm 8.25}$ & 0.10 & $58.64_{\pm 8.45}$ & 1.70 & $67.00_{\pm 9.26}$ & 0.90 & $79.00_{\pm 8.02}$ \\
 & 15       & 0.20 & $33.78_{\pm 8.48}$ & 0.30 & $36.86_{\pm 7.75}$ & 0.10 & $55.05_{\pm 8.57}$ & 1.80 & $67.00_{\pm 9.26}$ & 0.90 & $80.00_{\pm 7.88}$ \\
 & 20       & 0.20 & $38.66_{\pm 8.72}$ & 0.30 & $38.61_{\pm 5.85}$ & 0.10 & $58.46_{\pm 8.34}$ & 1.80 & $70.00_{\pm 9.03}$ & 0.90 & $79.00_{\pm 8.02}$ \\
 & 25       & 0.20 & $30.47_{\pm 8.12}$ & 0.30 & $36.39_{\pm 7.82}$ & 0.10 & $57.19_{\pm 8.49}$ & 1.80 & $66.00_{\pm 9.33}$ & 0.90 & $80.00_{\pm 7.84}$ \\
 & 30       & 0.20 & $34.21_{\pm 8.63}$ & 0.30 & $39.34_{\pm 8.00}$ & 0.10 & $59.00_{\pm 8.47}$ & 1.80 & $71.00_{\pm 8.94}$ & 0.90 & $77.00_{\pm 8.29}$ \\
 & $\infty$ & 0.20 & $34.19_{\pm 8.61}$ & 0.30 & $40.85_{\pm 8.09}$ & 0.10 & $56.86_{\pm 8.46}$ & 1.80 & $70.00_{\pm 9.03}$ & 0.90 & $83.00_{\pm 7.40}$ \\
\bottomrule
\end{tabular}}
\caption{\label{tab:overall_performance_qwen_ci} Overall Performance. We use Qwen3-8B as our backbone model. 
\textbf{C} and \textbf{F1} denote the average optimization cost (\texttt{avg\_cost}) and F1 score, respectively. Reported \textbf{F1} scores incorporate 95\% confidence intervals (CIs) formatted as $\text{Value}_{\pm \text{Margin}}$. For Top-k, \textbf{C} is marked \textsc{N/A} because this baseline ignores access costs during selection.
}
\end{table*}

%% file: table/overall_performance_llama_ci.tex
\begin{table*}[htbp]
\centering
\scriptsize
\setlength{\tabcolsep}{1.2pt}
\resizebox{\linewidth}{!}{
\begin{tabular}{ll cc cc cc cc cc}
\toprule
\multirow{3}{*}{\textbf{Method}} & \multirow{3}{*}{\textbf{Bgt/k}} 
& \multicolumn{6}{c}{\textbf{Open-Domain Datasets}} 
& \multicolumn{4}{c}{\textbf{Domain-Specific Datasets}} \\
\cmidrule(lr){3-8} \cmidrule(lr){9-12}
& & \multicolumn{2}{c}{\textbf{HotpotQA}} 
& \multicolumn{2}{c}{\textbf{NQ}} 
& \multicolumn{2}{c}{\textbf{TriviaQA}} 
& \multicolumn{2}{c}{\textbf{MedQA}} 
& \multicolumn{2}{c}{\textbf{MMLU}} \\
\cmidrule(lr){3-4} \cmidrule(lr){5-6} \cmidrule(lr){7-8} \cmidrule(lr){9-10} \cmidrule(lr){11-12}
& & \textbf{C} & \textbf{F1 (\%)} 
& \textbf{C} & \textbf{F1 (\%)} 
& \textbf{C} & \textbf{F1 (\%)} 
& \textbf{C} & \textbf{F1 (\%)} 
& \textbf{C} & \textbf{F1 (\%)} \\
\midrule
Vanilla LLM & - & \textsc{N/A} & $14.61_{\pm 4.63}$ & \textsc{N/A} & $12.52_{\pm 3.98}$ & \textsc{N/A} & $24.28_{\pm 5.82}$ & \textsc{N/A} & $55.00_{\pm 9.80}$ & \textsc{N/A}  & $67.00_{\pm 9.26}$ \\
\midrule
\multirow{6}{*}{Top-k} 
 & 5  & \textsc{N/A} &  $15.36_{\pm 5.55}$ & \textsc{N/A} & $27.55_{\pm 6.90}$ & \textsc{N/A}  & $47.41_{\pm 8.70}$ & \textsc{N/A} & $56.00_{\pm 9.78}$ & \textsc{N/A}  & $64.00_{\pm 9.46}$ \\
 & 10 & \textsc{N/A} &  $26.53_{\pm 7.65}$ & \textsc{N/A} & $33.55_{\pm 7.66}$ & \textsc{N/A}  & $54.38_{\pm 8.62}$ & \textsc{N/A} & $62.00_{\pm 9.56}$ & \textsc{N/A}  & $63.00_{\pm 9.51}$ \\
 & 15 & \textsc{N/A} &  $30.24_{\pm 7.98}$ & \textsc{N/A} & $36.57_{\pm 7.92}$ & \textsc{N/A}  & $59.13_{\pm 8.37}$ & \textsc{N/A} & $56.00_{\pm 9.78}$ & \textsc{N/A}  & $72.00_{\pm 8.84}$ \\
 & 20 & \textsc{N/A} &  $33.99_{\pm 8.30}$ & \textsc{N/A} & $34.52_{\pm 7.61}$ & \textsc{N/A}  & $63.23_{\pm 8.31}$ & \textsc{N/A} & $52.00_{\pm 9.84}$ & \textsc{N/A}  & $68.00_{\pm 9.19}$ \\
 & 25 & \textsc{N/A} &  $32.98_{\pm 8.24}$ & \textsc{N/A} & $39.60_{\pm 8.21}$ & \textsc{N/A}  & $59.45_{\pm 8.41}$ & \textsc{N/A} & $54.00_{\pm 9.82}$ & \textsc{N/A}  & $67.00_{\pm 9.26}$ \\
 & 30 & \textsc{N/A} &  $32.69_{\pm 8.18}$ & \textsc{N/A} & $33.33_{\pm 7.53}$ & \textsc{N/A}  & $60.40_{\pm 8.47}$ & \textsc{N/A} & $59.00_{\pm 9.69}$ & \textsc{N/A}  & $71.00_{\pm 8.94}$ \\
\midrule
\multirow{8}{*}{Relevance} 
 & 0        & 0.00 & $33.45_{\pm 8.20}$ & 0.00 & $39.48_{\pm 8.10}$ & 0.00 & $61.19_{\pm 8.40}$ & 0.00 & $56.00_{\pm 9.78}$ & 0.00 & $61.00_{\pm 9.61}$ \\
 & 5        & 4.80 & $31.23_{\pm 8.01}$ & 5.00 & $33.90_{\pm 7.79}$ & 4.90 & $60.87_{\pm 8.28}$ & 5.00 & $57.00_{\pm 9.75}$ & 5.00 & $62.00_{\pm 9.56}$ \\
 & 10       & 9.20 & $34.47_{\pm 8.52}$ & 9.70 & $37.12_{\pm 7.97}$ & 9.10 & $59.18_{\pm 8.48}$ & 10.0 & $53.00_{\pm 9.83}$ & 9.90 & $66.00_{\pm 9.33}$ \\
 & 15       & 12.9 & $36.99_{\pm 8.56}$ & 13.6 & $42.44_{\pm 8.25}$ & 12.6 & $60.93_{\pm 8.35}$ & 14.9 & $56.00_{\pm 9.78}$ & 14.7 & $71.00_{\pm 8.94}$ \\
 & 20       & 15.5 & $41.02_{\pm 8.69}$ & 16.8 & $41.36_{\pm 8.00}$ & 15.6 & $60.29_{\pm 8.40}$ & 19.9 & $47.00_{\pm 9.83}$ & 19.6 & $62.00_{\pm 9.56}$ \\
 & 25       & 17.2 & $33.55_{\pm 8.27}$ & 19.2 & $31.93_{\pm 7.76}$ & 18.0 & $64.47_{\pm 8.08}$ & 25.0 & $52.00_{\pm 9.84}$ & 24.1 & $70.00_{\pm 9.11}$ \\
 & 30       & 18.1 & $37.73_{\pm 8.55}$ & 20.9 & $39.51_{\pm 7.97}$ & 19.7 & $60.62_{\pm 8.34}$ & 29.8 & $53.00_{\pm 9.83}$ & 28.1 & $65.00_{\pm 9.40}$ \\
 & $\infty$ & 19.2 & $37.14_{\pm 8.46}$ & 23.5 & $34.71_{\pm 7.86}$ & 22.3 & $61.81_{\pm 8.20}$ & 61.7 & $55.00_{\pm 9.80}$ & 48.2 & $66.00_{\pm 9.33}$ \\
\midrule
\multirow{8}{*}{Greedy} 
 & 0        & 0.00 & $30.48_{\pm 8.10}$ & 0.00 & $35.42_{\pm 8.18}$ & 0.00 & $58.03_{\pm 8.23}$ & 0.00 & $53.00_{\pm 9.83}$ & 0.00 & $67.00_{\pm 9.26}$ \\
 & 5        & 4.74 & $31.35_{\pm 7.96}$ & 4.89 & $34.25_{\pm 7.81}$ & 4.82 & $56.84_{\pm 8.34}$ & 4.78 & $51.00_{\pm 9.85}$ & 4.82 & $58.00_{\pm 9.72}$ \\
 & 10       & 8.93 & $34.21_{\pm 8.33}$ & 9.40 & $34.43_{\pm 7.99}$ & 8.78 & $59.36_{\pm 8.41}$ & 9.22 & $48.00_{\pm 9.84}$ & 9.13 & $67.00_{\pm 9.26}$ \\
 & 15       & 12.56& $36.26_{\pm 8.60}$ & 12.94& $36.34_{\pm 8.17}$ & 12.03& $61.44_{\pm 8.36}$ & 13.73& $57.00_{\pm 9.75}$ & 13.68& $67.00_{\pm 9.26}$ \\
 & 20       & 15.08& $36.19_{\pm 8.44}$ & 16.10& $36.87_{\pm 7.96}$ & 14.94& $62.12_{\pm 8.31}$ & 18.54& $54.00_{\pm 9.82}$ & 18.16& $70.00_{\pm 9.11}$ \\
 & 25       & 16.98& $35.17_{\pm 8.47}$ & 18.72& $34.17_{\pm 7.94}$ & 17.60& $61.18_{\pm 8.27}$ & 23.50& $50.00_{\pm 9.80}$ & 22.99& $76.00_{\pm 8.41}$ \\
 & 30       & 18.00& $37.80_{\pm 8.49}$ & 20.54& $36.66_{\pm 8.15}$ & 19.38& $63.71_{\pm 8.32}$ & 28.57& $55.00_{\pm 9.80}$ & 27.08& $71.00_{\pm 8.94}$ \\
 & $\infty$ & 19.24& $35.74_{\pm 8.46}$ & 23.46& $33.60_{\pm 7.83}$ & 22.30& $60.34_{\pm 8.33}$ & 61.73& $50.00_{\pm 9.80}$ & 48.24& $75.00_{\pm 9.26}$ \\
\midrule
\multirow{8}{*}{Knapsack} 
 & 0        & 0.00 & $33.91_{\pm 8.15}$ & 0.00 & $32.21_{\pm 7.89}$ & 0.00 & $56.13_{\pm 8.16}$ & 0.00 & $51.00_{\pm 9.85}$ & 0.00 & $60.00_{\pm 9.60}$ \\
 & 5        & 4.77 & $33.56_{\pm 8.15}$ & 4.97 & $35.91_{\pm 7.91}$ & 4.82 & $58.87_{\pm 8.41}$ & 4.87 & $54.00_{\pm 9.75}$ & 4.88 & $64.00_{\pm 9.46}$ \\
 & 10       & 9.17 & $33.05_{\pm 8.20}$ & 9.46 & $34.52_{\pm 7.80}$ & 8.87 & $58.50_{\pm 8.46}$ & 9.58 & $53.00_{\pm 9.83}$ & 9.55 & $57.00_{\pm 9.75}$ \\
 & 15       & 12.65& $36.62_{\pm 8.56}$ & 13.21& $35.70_{\pm 7.83}$ & 12.30& $64.00_{\pm 8.32}$ & 14.30& $58.00_{\pm 9.72}$ & 14.10& $69.00_{\pm 9.11}$ \\
 & 20       & 15.38& $33.59_{\pm 8.40}$ & 16.49& $36.88_{\pm 7.90}$ & 15.45& $60.05_{\pm 8.39}$ & 19.20& $51.00_{\pm 9.85}$ & 18.97& $72.00_{\pm 8.84}$ \\
 & 25       & 17.10& $36.65_{\pm 8.53}$ & 18.94& $33.35_{\pm 7.73}$ & 17.81& $61.69_{\pm 8.24}$ & 24.31& $57.00_{\pm 9.75}$ & 23.59& $69.00_{\pm 9.11}$ \\
 & 30       & 18.06& $34.43_{\pm 8.49}$ & 20.72& $37.65_{\pm 8.06}$ & 19.59& $62.29_{\pm 8.27}$ & 29.26& $52.00_{\pm 9.84}$ & 27.68& $79.00_{\pm 8.02}$ \\
 & $\infty$ & 19.24& $37.00_{\pm 8.54}$ & 23.46& $35.31_{\pm 7.91}$ & 22.30& $62.57_{\pm 8.29}$ & 61.73& $54.00_{\pm 9.82}$ & 48.24& $72.00_{\pm 8.84}$ \\
\midrule
\multirow{8}{*}{MMR} 
 & 0        & 0.00 & $29.69_{\pm 7.69}$  & 0.00 & $33.52_{\pm 8.08}$  & 0.00 &  $56.99_{\pm 8.41}$  & 0.00 & $47.00_{\pm 9.83}$ & 0.00 & $55.00_{\pm 9.80}$ \\
 & 5        & 4.67 & $28.78_{\pm 7.89}$  & 4.86 & $33.25_{\pm 7.88}$  & 4.73 &  $60.37_{\pm 8.39}$  & 4.66 & $52.00_{\pm 9.84}$ & 4.62 & $70.00_{\pm 9.11}$ \\
 & 10       & 8.04 & $34.52_{\pm 8.23}$  & 8.78 & $34.21_{\pm 7.99}$  & 8.00 &  $62.89_{\pm 8.13}$  & 7.58 & $56.00_{\pm 9.75}$ & 7.85 & $69.00_{\pm 9.11}$ \\
 & 15       & 9.82 & $35.19_{\pm 8.23}$  & 11.12& $34.70_{\pm 8.16}$  & 9.87 &  $63.78_{\pm 8.38}$  & 8.81 & $53.00_{\pm 9.83}$ & 9.12 & $66.00_{\pm 9.33}$ \\
 & 20       & 10.61& $33.51_{\pm 8.27}$  & 12.04& $36.45_{\pm 8.11}$  & 10.86&  $61.92_{\pm 8.32}$  & 9.18 & $53.00_{\pm 9.83}$ & 9.52 & $63.00_{\pm 9.51}$ \\
 & 25       & 10.88& $38.97_{\pm 8.58}$  & 12.46& $33.31_{\pm 7.66}$  & 11.16&  $61.07_{\pm 8.29}$  & 9.34 & $58.00_{\pm 9.75}$ & 9.63 & $66.00_{\pm 9.33}$ \\
 & 30       & 10.88& $34.68_{\pm 8.33}$  & 12.56& $36.55_{\pm 8.26}$  & 11.18&  $63.55_{\pm 8.38}$  & 9.37 & $53.00_{\pm 9.83}$ & 9.68 & $65.00_{\pm 9.40}$ \\
 & $\infty$ & 10.88& $35.69_{\pm 8.43}$  & 12.56& $38.33_{\pm 8.25}$  & 11.18&  $61.83_{\pm 8.32}$  & 9.37 & $58.00_{\pm 9.72}$ & 9.68 & $66.00_{\pm 9.33}$ \\
\midrule
\multirow{8}{*}{Redundancy} 
 & 0        & 0.00 & $30.62_{\pm 8.01}$  & 0.00 & $32.14_{\pm 7.91}$  & 0.00 & $58.66_{\pm 8.27}$  & 0.00 & $47.00_{\pm 9.83}$ & 0.00 & $68.00_{\pm 9.26}$ \\
 & 5        & 0.15 & $30.36_{\pm 7.96}$  & 0.26 & $34.12_{\pm 7.86}$  & 0.09 & $56.26_{\pm 8.46}$  & 1.18 & $52.00_{\pm 9.84}$ & 0.64 & $59.00_{\pm 9.69}$ \\
 & 10       & 0.19 & $30.46_{\pm 8.00}$  & 0.30 & $35.01_{\pm 8.06}$  & 0.09 & $57.41_{\pm 8.47}$  & 1.68 & $59.00_{\pm 9.11}$ & 0.91 & $64.00_{\pm 9.40}$ \\
 & 15       & 0.19 & $30.71_{\pm 8.11}$  & 0.30 & $34.70_{\pm 8.11}$  & 0.09 & $59.05_{\pm 8.52}$  & 1.76 & $57.00_{\pm 9.75}$ & 0.91 & $66.00_{\pm 9.33}$ \\
 & 20       & 0.19 & $32.29_{\pm 8.06}$  & 0.30 & $36.23_{\pm 8.06}$  & 0.09 & $57.42_{\pm 8.46}$  & 1.76 & $54.00_{\pm 9.82}$ & 0.91 & $64.00_{\pm 9.40}$ \\
 & 25       & 0.19 & $28.53_{\pm 7.73}$  & 0.30 & $34.87_{\pm 8.06}$  & 0.09 & $57.60_{\pm 8.49}$  & 1.76 & $59.00_{\pm 9.11}$ & 0.91 & $65.00_{\pm 9.40}$ \\
 & 30       & 0.19 & $28.74_{\pm 7.86}$  & 0.30 & $36.23_{\pm 8.15}$  & 0.09 & $58.24_{\pm 8.49}$  & 1.76 & $51.00_{\pm 9.84}$ & 0.91 & $63.00_{\pm 9.51}$ \\
 & $\infty$ & 0.19 & $33.32_{\pm 8.43}$  & 0.30 & $35.73_{\pm 7.91}$  & 0.09 & $57.66_{\pm 8.41}$  & 1.76 & $51.00_{\pm 9.84}$ & 1.76 & $64.00_{\pm 9.40}$ \\
\bottomrule
\end{tabular}}
\caption{\label{tab:overall_performance_llama_ci}Overall Performance. We use Llama3.1-8B-instruct as our backbone model. \textbf{C} and \textbf{F1} denote the average optimization cost (\texttt{avg\_cost}) and F1 score, respectively. Reported \textbf{F1} scores incorporate 95\% confidence intervals (CIs) formatted as $\text{Value}_{\pm \text{Margin}}$. For Top-k, \textbf{C} is marked \textsc{N/A} because this baseline ignores access costs during selection.}
\end{table*}

%% file: table/overall_costmore_ci.tex
\begin{table*}[htbp]
\centering
\scriptsize
\setlength{\tabcolsep}{1.2pt}
\resizebox{\linewidth}{!}{
\begin{tabular}{ll cc cc cc cc cc}
\toprule
\multirow{3}{*}{\textbf{Method}} & \multirow{3}{*}{\textbf{Bgt/k}} 
& \multicolumn{6}{c}{\textbf{Open-Domain Datasets}} 
& \multicolumn{4}{c}{\textbf{Domain-Specific Datasets}} \\
\cmidrule(lr){3-8} \cmidrule(lr){9-12}
& & \multicolumn{2}{c}{\textbf{HotpotQA}} 
& \multicolumn{2}{c}{\textbf{NQ}} 
& \multicolumn{2}{c}{\textbf{TriviaQA}} 
& \multicolumn{2}{c}{\textbf{MedQA}} 
& \multicolumn{2}{c}{\textbf{MMLU}} \\
\cmidrule(lr){3-4} \cmidrule(lr){5-6} \cmidrule(lr){7-8} \cmidrule(lr){9-10} \cmidrule(lr){11-12}
& & \textbf{C} & \textbf{F1 (\%)} 
& \textbf{C} & \textbf{F1 (\%)} 
& \textbf{C} & \textbf{F1 (\%)} 
& \textbf{C} & \textbf{F1 (\%)} 
& \textbf{C} & \textbf{F1 (\%)} \\
\midrule
Vanilla LLM & - & \textsc{N/A} & $39.10_{\pm 8.71}$ & \textsc{N/A} & $21.89_{\pm 6.74}$ & \textsc{N/A} & $41.49_{\pm 8.67}$ & \textsc{N/A} & $44.00_{\pm 9.78}$ & \textsc{N/A} & $74.00_{\pm 8.64}$ \\
\midrule
\multirow{6}{*}{Top-k} 
 & 5  & \textsc{N/A} & $34.14_{\pm 8.59}$ & \textsc{N/A} & $35.61_{\pm 7.50}$ & \textsc{N/A} & $56.85_{\pm 8.75}$ & \textsc{N/A} & $71.00_{\pm 8.94}$ & \textsc{N/A} & $81.00_{\pm 7.69}$ \\
 & 10 & \textsc{N/A} & $32.75_{\pm 8.48}$ & \textsc{N/A} & $37.91_{\pm 7.72}$ & \textsc{N/A} & $54.74_{\pm 8.73}$ & \textsc{N/A} & $71.00_{\pm 8.94}$ & \textsc{N/A} & $80.00_{\pm 7.84}$ \\
 & 15 & \textsc{N/A} & $36.07_{\pm 8.70}$ & \textsc{N/A} & $39.97_{\pm 8.26}$ & \textsc{N/A} & $58.97_{\pm 8.49}$ & \textsc{N/A} & $65.00_{\pm 9.40}$ & \textsc{N/A} & $79.00_{\pm 7.98}$ \\
 & 20 & \textsc{N/A} & $34.53_{\pm 8.63}$ & \textsc{N/A} & $38.55_{\pm 8.23}$ & \textsc{N/A} & $59.92_{\pm 8.42}$ & \textsc{N/A} & $61.00_{\pm 9.56}$ & \textsc{N/A} & $82.00_{\pm 7.47}$ \\
 & 25 & \textsc{N/A} & $32.98_{\pm 8.24}$ & \textsc{N/A} & $40.37_{\pm 8.02}$ & \textsc{N/A} & $58.88_{\pm 8.61}$ & \textsc{N/A} & $62.00_{\pm 9.56}$ & \textsc{N/A} & $82.00_{\pm 7.47}$ \\
 & 30 & \textsc{N/A} & $37.45_{\pm 8.90}$ & \textsc{N/A} & $35.12_{\pm 7.64}$ & \textsc{N/A} & $63.49_{\pm 8.31}$ & \textsc{N/A} & $61.00_{\pm 9.56}$ & \textsc{N/A} & $79.00_{\pm 7.98}$ \\
\midrule
\multirow{8}{*}{Relevance} 
 & 0        & 0.00 & $36.91_{\pm 8.73}$ & 0.00 & $39.48_{\pm 8.34}$ & 0.00 & $55.38_{\pm 8.52}$  & 0.00 &  $62.00_{\pm 9.56}$ & 0.00 & $77.00_{\pm 8.25}$ \\
 & 5        & 4.84 & $34.71_{\pm 8.63}$ & 4.98 & $44.19_{\pm 8.36}$ & 4.87 & $56.02_{\pm 8.66}$  & 5.00 &  $66.00_{\pm 9.33}$ & 4.98 & $84.00_{\pm 7.22}$ \\
 & 10       & 9.10 & $37.89_{\pm 8.91}$ & 9.51 & $41.44_{\pm 8.08}$ & 8.97 & $57.51_{\pm 8.78}$  & 9.99 &  $63.00_{\pm 9.51}$ & 9.95 & $80.00_{\pm 8.00}$ \\
 & 15       & 12.35& $36.15_{\pm 8.85}$ & 13.29& $42.44_{\pm 8.34}$ & 12.26& $60.77_{\pm 8.63}$  & 15.00&  $65.00_{\pm 9.40}$ & 14.61& $80.00_{\pm 8.00}$ \\
 & 20       & 14.29& $35.79_{\pm 8.53}$ & 15.85& $41.36_{\pm 8.21}$ & 14.85& $60.14_{\pm 8.55}$  & 19.88&  $62.00_{\pm 9.56}$ & 18.96& $80.00_{\pm 8.00}$ \\
 & 25       & 15.01& $35.28_{\pm 8.69}$ & 17.24& $40.21_{\pm 8.13}$ & 16.24& $59.96_{\pm 8.46}$  & 24.57&  $63.00_{\pm 9.51}$ & 22.63& $82.00_{\pm 7.47}$ \\
 & 30       & 15.26& $37.79_{\pm 8.97}$ & 17.83& $39.51_{\pm 7.97}$ & 16.62& $60.62_{\pm 8.34}$  & 28.79&  $62.00_{\pm 9.56}$ & 25.50& $80.00_{\pm 7.84}$ \\
 & $\infty$ & 15.40& $36.68_{\pm 8.74}$ & 17.98& $38.63_{\pm 8.13}$ & 16.80& $62.92_{\pm 8.36}$  & 35.55&  $68.00_{\pm 9.19}$ & 28.96& $77.00_{\pm 8.29}$ \\
\midrule
\multirow{8}{*}{Greedy} 
 & 0        &  0.00 & $31.09_{\pm 8.16}$ & 0.00 &  $37.12_{\pm 7.95}$  & 0.00 & $55.64_{\pm 8.48}$  & 0.00 &  $59.00_{\pm 8.87}$  & 0.00 &  $71.00_{\pm 8.94}$ \\
 & 5        &  4.84 & $32.90_{\pm 8.37}$ & 4.95 &  $40.77_{\pm 8.31}$  & 4.82 & $58.01_{\pm 8.63}$  & 28.49&  $55.00_{\pm 9.80}$  & 25.24&  $71.00_{\pm 8.94}$ \\
 & 10       &  8.99 & $40.66_{\pm 8.86}$ & 9.41 &  $39.89_{\pm 8.09}$  & 8.84 & $58.93_{\pm 8.71}$  & 9.37 &  $64.00_{\pm 9.46}$  & 9.68 &  $80.00_{\pm 7.84}$ \\
 & 15       &  12.27& $36.65_{\pm 8.82}$ & 13.06&  $38.46_{\pm 8.24}$  & 12.15& $61.30_{\pm 8.54}$  & 1.49 &  $66.00_{\pm 9.33}$  & 0.70 &  $78.00_{\pm 8.16}$ \\
 & 20       &  14.22& $35.91_{\pm 8.69}$ & 15.66&  $40.71_{\pm 8.02}$  & 14.68& $60.32_{\pm 8.48}$  & 28.79&  $62.00_{\pm 9.56}$  & 25.50&  $80.00_{\pm 7.84}$ \\
 & 25       &  15.00& $37.77_{\pm 8.88}$ & 17.18&  $40.43_{\pm 8.11}$  & 16.20& $60.58_{\pm 8.41}$  & 1.49 &  $68.00_{\pm 9.11}$  & 0.70 &  $79.00_{\pm 8.02}$ \\
 & 30       &  15.24& $37.80_{\pm 8.49}$ & 17.80&  $39.83_{\pm 8.29}$  & 16.60& $63.71_{\pm 8.41}$  & 28.79&  $53.00_{\pm 9.83}$  & 25.50&  $80.00_{\pm 7.84}$ \\
 & $\infty$ &  15.40& $37.68_{\pm 8.73}$ & 17.98&  $38.36_{\pm 7.83}$  & 16.80& $65.70_{\pm 8.19}$  & 35.55&  $65.00_{\pm 9.40}$  & 28.96&  $75.00_{\pm 8.53}$ \\
\midrule
\multirow{8}{*}{Knapsack} 
 & 0        & 0.00 & $37.37_{\pm 8.79}$ & 0.00 & $38.78_{\pm 8.10}$ & 0.00 & $59.37_{\pm 8.47}$ & 0.00 & $65.00_{\pm 9.40}$ & 0.00 & $79.00_{\pm 8.02}$ \\
 & 5        & 4.84 & $36.85_{\pm 8.67}$ & 4.98 & $41.53_{\pm 8.23}$ & 4.87 & $59.41_{\pm 8.56}$ & 29.26& $64.00_{\pm 9.46}$ & 27.68& $79.00_{\pm 7.98}$ \\
 & 10       & 9.10 & $38.67_{\pm 8.75}$ & 9.50 & $45.44_{\pm 8.34}$ & 8.97 & $59.66_{\pm 8.64}$ & 1.49 & $66.00_{\pm 9.33}$ & 0.70 & $78.00_{\pm 8.16}$ \\
 & 15       & 12.34& $37.74_{\pm 8.78}$ & 13.21& $41.46_{\pm 8.33}$ & 12.26& $62.29_{\pm 8.33}$ & 35.55& $65.00_{\pm 9.40}$ & 28.96& $77.00_{\pm 8.20}$ \\
 & 20       & 14.27& $40.00_{\pm 8.98}$ & 15.80& $38.80_{\pm 7.89}$ & 14.85& $62.61_{\pm 8.46}$ & 1.49 & $64.00_{\pm 9.46}$ & 0.70 & $79.00_{\pm 8.02}$ \\
 & 25       & 15.01& $36.94_{\pm 8.78}$ & 17.24& $41.94_{\pm 7.96}$ & 16.24& $61.67_{\pm 8.34}$ & 35.55& $64.00_{\pm 9.46}$ & 28.96& $72.00_{\pm 8.84}$ \\
 & 30       & 15.26& $37.98_{\pm 8.84}$ & 17.83& $37.86_{\pm 7.86}$ & 16.62& $62.73_{\pm 8.33}$ & 1.49 & $65.00_{\pm 9.40}$ & 0.70 & $79.00_{\pm 7.98}$ \\
 & $\infty$ & 15.40& $37.66_{\pm 8.91}$ & 23.46& $38.95_{\pm 8.00}$ & 16.80& $62.57_{\pm 8.29}$ & 35.55& $61.00_{\pm 9.61}$ & 28.96& $79.00_{\pm 7.98}$ \\
\midrule
\multirow{8}{*}{MMR} 
 & 0        & 0.00 & $35.10_{\pm 8.40}$ & 0.00 & $38.46_{\pm 8.08}$ & 0.00 & $57.09_{\pm 8.51}$ & 0.00 & $63.00_{\pm 9.51}$ & 0.00 & $82.00_{\pm 7.47}$ \\
 & 5        & 4.67 & $34.85_{\pm 8.47}$ & 4.92 & $43.58_{\pm 8.10}$ & 4.80 & $59.54_{\pm 8.49}$ & 4.93 & $64.00_{\pm 9.46}$ & 4.92 & $83.00_{\pm 7.40}$ \\
 & 10       & 8.04 & $37.83_{\pm 8.77}$ & 9.39 & $43.40_{\pm 8.26}$ & 8.60 & $62.89_{\pm 8.33}$ & 9.75 & $65.00_{\pm 9.40}$ & 9.65 & $82.00_{\pm 7.47}$ \\
 & 15       & 9.82 & $39.54_{\pm 8.87}$ & 12.56& $42.35_{\pm 8.16}$ & 11.18& $62.18_{\pm 8.32}$ & 27.59& $64.00_{\pm 9.46}$ & 23.00& $80.00_{\pm 7.84}$ \\
 & 20       & 11.38& $38.49_{\pm 8.55}$ & 12.81& $42.35_{\pm 8.00}$ & 11.28& $62.18_{\pm 8.30}$ & 28.79& $65.00_{\pm 9.40}$ & 25.50& $80.00_{\pm 7.84}$ \\
 & 25       & 11.77& $36.06_{\pm 8.71}$ & 15.97& $40.64_{\pm 7.98}$ & 13.74& $61.07_{\pm 8.29}$ & 28.79& $67.00_{\pm 9.26}$ & 25.50& $80.00_{\pm 7.84}$ \\
 & 30       & 11.77& $37.68_{\pm 8.84}$ & 16.34& $42.35_{\pm 8.33}$ & 13.86& $63.55_{\pm 8.38}$ & 28.79& $65.00_{\pm 9.40}$ & 25.50& $80.00_{\pm 7.84}$ \\
 & $\infty$ & 11.77& $36.30_{\pm 8.84}$ & 16.42& $43.86_{\pm 8.25}$ & 13.90& $61.83_{\pm 8.32}$ & 35.55& $62.00_{\pm 9.56}$ & 28.96& $81.00_{\pm 7.47}$ \\
\midrule
\multirow{8}{*}{Redundancy} 
 & 0        & 0.00 & $36.61_{\pm 8.62}$ & 0.00 & $36.97_{\pm 7.91}$ & 0.00 & $55.60_{\pm 8.53}$  & 0.00 & $62.00_{\pm 9.56}$ & 0.00 & $82.00_{\pm 7.57}$ \\
 & 5        & 0.15 & $34.16_{\pm 8.66}$ & 0.26 & $38.93_{\pm 8.11}$ & 0.09 & $57.39_{\pm 8.47}$  & 1.49 & $67.00_{\pm 9.26}$ & 0.70 & $80.00_{\pm 7.88}$ \\
 & 10       & 0.19 & $37.80_{\pm 8.84}$ & 0.30 & $37.86_{\pm 7.86}$ & 0.09 & $59.37_{\pm 8.47}$  & 1.49 & $65.00_{\pm 9.40}$ & 0.70 & $79.00_{\pm 7.98}$ \\
 & 15       & 0.19 & $37.98_{\pm 8.84}$ & 0.30 & $37.65_{\pm 8.06}$ & 0.09 & $62.29_{\pm 8.27}$  & 1.49 & $65.00_{\pm 9.40}$ & 0.70 & $79.00_{\pm 7.98}$ \\
 & 20       & 0.19 & $37.68_{\pm 8.84}$ & 0.30 & $39.17_{\pm 8.06}$ & 0.09 & $58.24_{\pm 8.47}$  & 1.49 & $66.00_{\pm 9.33}$ & 0.70 & $78.00_{\pm 8.16}$ \\
 & 25       & 0.19 & $34.16_{\pm 8.66}$ & 0.30 & $41.96_{\pm 8.06}$ & 0.09 & $59.96_{\pm 8.46}$  & 1.49 & $68.00_{\pm 9.19}$ & 0.70 & $80.00_{\pm 7.84}$ \\
 & 30       & 0.19 & $37.79_{\pm 8.97}$ & 0.30 & $39.51_{\pm 7.97}$ & 0.09 & $56.27_{\pm 8.49}$  & 1.49 & $66.00_{\pm 9.33}$ & 0.70 & $78.00_{\pm 8.16}$ \\
 & $\infty$ & 0.19 & $33.32_{\pm 8.49}$ & 0.30 & $36.23_{\pm 8.06}$ & 0.09 & $57.66_{\pm 8.41}$  & 1.49 & $66.00_{\pm 9.33}$ & 1.49 & $79.00_{\pm 8.02}$ \\
\bottomrule
\end{tabular}}
\caption{\label{tab:overall_performance_qwen_costmore_ci}Overall Performance when Pricing Setting is 0, 1, and 2 for Tier 0, 1, and 2, respectively. We use Qwen3-8B as our backbone model. \textbf{C} and \textbf{F1} denote average optimization cost (\texttt{avg\_cost}) and F1 score, respectively. Reported \textbf{F1} scores incorporate 95\% confidence intervals (CIs) formatted as $\text{Value}_{\pm \text{Margin}}$. For Top-k, \textbf{C} is marked \textsc{N/A} because this baseline ignores access costs during selection.}
\end{table*}